\documentclass[journal]{IEEEtran}
\usepackage{cite}
\usepackage{amsmath,amssymb,amsfonts}
\usepackage{algorithm}
\usepackage{algorithmic}
\usepackage{graphicx}
\usepackage{multirow}
\usepackage{textcomp}
\usepackage{xcolor}
\usepackage{graphicx}
\usepackage{subfigure}
\usepackage{CJKutf8}
\usepackage{etoolbox}
\usepackage[T1]{fontenc}
\newenvironment{sloppypar*}

\begin{document}

\title{Efficient Federated Learning with Spike Neural Networks for Traffic Sign Recognition}

\author{ Kan Xie,  Zhe Zhang,  Bo Li, Jiawen Kang, Dusit Niyato,~\IEEEmembership{Fellow,~IEEE},  Shengli Xie,~\IEEEmembership{Fellow,~IEEE}, Yi Wu

\thanks{This research is supported, in part, by the programme DesCartes and is supported by the National Research Foundation, Prime Minister’s Office, Singapore under its Campus for Research Excellence and Technological Enterprise (CREATE) programme, Alibaba Group through Alibaba Innovative Research (AIR) Program and Alibaba-NTU Singapore Joint Research Institute (JRI), the National Research Foundation, Singapore under the AI Singapore Programme (AISG) (AISG2-RP-2020-019), and Singapore Ministry of Education (MOE) Tier 1 (RG16/20). (\textit{Corresponding author: Jiawen Kang.})}
\thanks{Copyright (c) 2015 IEEE. Personal use of this material is permitted. However, permission to use this material for any other purposes must be obtained from the IEEE by sending a request to pubs-permissions@ieee.org.}
\thanks{Kan Xie is with the School of Automation, Guangdong University of Technology, Guangzhou 510006, China, and also with the Key Laboratory of Intelligent Information Processing and System Integration of IoT, Ministry of Education, Guangzhou 510006, China (e-mail: kxie@gdut.edu.cn).}
\thanks{Bo Li is with the School of Automation, Guangdong University of Technology, Guangzhou 510006, China, and also with the Guangdong-HongKong-Macao Joint Laboratory for Smart Discrete Manufacturing, Guangzhou 510006, China (e-mail: libo999@gdut.edu.cn).}
\thanks{Jiawen Kang is with the School of Automation, Guangdong University of Technology, Guangzhou 510006, China, and also with the 111 Center for Intelligent Batch Manufacturing Based on IoT Technology, Guangzhou 510006, China (e-mail: kavinkang@gdut.edu.cn).}
\thanks{Shengli Xie is with School of Automation, Guangdong University of Technology, Guangzhou 510006, China, and also with the Guangdong Key Laboratory of IoT Information Technology, Guangzhou 510006, China (e-mail: shlxie@gdut.edu.cn).}
\thanks{Z. Zhang and Y. Wu are with the School of Data Science and Technology, Heilongjiang University, Harbin 150080, China, and Y. Wu is also with the Institute for Cryptology \& Network Security, Heilongjiang University, Harbin 150080, China (e-mail: 2202517@s.hlju.edu.cn, 1995050@hlju.edu.cn).}
\thanks{D. Niyato is with School of Computer Science and Engineering,
Nanyang Technological University (e-mail: DNIYATO@ntu.edu.sg).}

}

%\markboth{Journal of \LaTeX\ Class Files,~Vol.~14, No.~8, August~2015}% {Shell \MakeLowercase{\textit{et al.}}: Bare Demo of IEEEtran.cls for IEEE Journals}

\maketitle

\begin{abstract}
With the gradual popularization of self-driving, it is becoming increasingly important for vehicles to smartly make the right driving decisions and autonomously obey traffic rules by correctly recognizing traffic signs. However, for machine learning-based traffic sign recognition on the Internet of Vehicles (IoV), a large amount of traffic sign data from distributed vehicles is needed to be gathered in a centralized server for model training, which brings serious privacy leakage risk because of traffic sign data containing lots of location privacy information. To address this issue, we first exploit privacy-preserving federated learning to perform collaborative training for accurate recognition models without sharing raw traffic sign data.
Nevertheless, due to the limited computing and energy resources of most devices, it is hard for vehicles to continuously undertake complex artificial intelligence tasks.
Therefore, we introduce powerful Spike Neural Networks (SNNs) into traffic sign recognition for energy-efficient and fast model training, which is the next generation of neural networks and is practical and well-fitted to IoV scenarios. 
{Furthermore, we design a novel encoding scheme for SNNs based on neuron receptive fields to extract information from the pixel and spatial dimensions of traffic signs to achieve high-accuracy training.
\textcolor{black}{Numerical results indicate that the proposed federated SNN outperforms traditional federated convolutional neural networks in terms of accuracy, noise immunity, and energy efficiency as well.}}
\end{abstract}


\section{Introduction}
According to the forecast report from well-known analysis organization Gartner, by 2023, there are more than 740,000 autonomous-ready vehicles added all around the world \cite{report}. It is obvious that autonomous driving is becoming more and more popular and shows huge potential to realize high-level intelligent transportation \cite{9771336}. When driving on the roads, traffic signs are detected and recognized by vehicles with powerful sensors and cameras. These signs play important roles to guide vehicles in an orderly manner and also improving driving safety, which provides significant assistance for vehicle control without interruption \cite{9366426}.
For example, warning signs remind vehicles to avoid obstacles and be aware of dangerous bends. The indicator signs can help the vehicles to perform pre-processing control and to follow road instructions, e.g, the speed limit on the highway. Therefore, the correct recognition and precise application of traffic signs can provide more perfect assistance for driver assistance systems and even autonomous driving \cite{9647926}.

Traditionally, to achieve precious and intelligent traffic sign recognition, the traffic signs are collected by vehicles and then gathered in a centralized server for machine learning-based analysis and recognition \cite{9016391}. However, these existing schemes have overlooked the privacy concerns of vehicles (e.g., some sensitive location information mapping to specific traffic signs on a particular location), which hinders data gathering and the training collaboration among vehicles \cite{liu2021vehicle}. Motivated by the advantages of privacy protection and breaking the data barrier, Federated Learning (FL) has been introduced into various vehicular applications \cite{liu2020privacy,9492053} and IoV scenarios \cite{8832210,lu2020blockchain}. Here, federated learning is a promising privacy-preserving machine learning paradigm that allows numerous vehicles to train a globally shared recognition model and only share a local training model without revealing individual raw data (i.e., collected traffic sign pictures) \cite{8994206,9562748}. 

For traffic sign recognition in IoV, due to limited energy resources and intermittent communications of vehicles, the energy efficiency, scalability, and robustness of model training need to be further enhanced for wide deployment \cite{8919978}. However, traditional FL methods pose challenges to the energy overhead and scalability with the rapid increase of vehicles in IoV. In addition, there exist blurred sensing traffic sign pictures caused by high-speed mobility of vehicles or bad weather conditions, it is increasingly important to improve the robustness of traffic sign recognition in IoV. 
All these issues inspire us to propose a new FL framework in this paper, where the emerging Spike Neural Network (SNN) technology is naturally integrated into IoV systems.
The SNNs, which are 3rd generation artificial neural networks inspired by information processing in biology, have been introduced to achieve efficient deep neural network implementation with favorable properties including low power consumption, fast inference, good robustness, and event-driven information processing \cite{pfeiffer2018deep}. Therefore, SNNs are a promising choice and highly suitable for training many machine learning tasks in IoV, especially traffic sign recognition  \cite{skatchkovsky2020federated}. To this end, as shown in Fig.~\ref{Model}, we design an efficient SNN-empowered federated learning framework for low-power and privacy-preserving traffic sign recognition.
\textcolor{black}{In each communication round, the networked vehicle uses the local traffic sign dataset to train the SNN model, which is then uploaded to a nearby Road Side Unit (RSU) for aggregation to obtain the global model of the next round.}
{However, this framework still faces a challenge in encoding traffic sign images for efficient SNNs.}

In SNNs, the training data needs to be encoded as a set of spike sequences with time steps.
Rate-based encoding \cite{6789203} has been widely used such as in \cite{wu2019direct,sengupta2019going,diehl2015unsupervised}, which utilizes the pixel intensities of images to extract input features.
The greater the intensity of a single pixel of an image, the more informative the spike sequence it encodes.
This method generally requires very high pixel values to be encoded into a sequence of high-frequency spikes for quick forward propagation in SNNs.
{Thus, rate-based coding has limitations when facing blurry traffic sign images captured by vehicles in bad weather conditions.}
Inspired by the biological vision, we find that the degree of stimulation of an image on organisms is not only affected by the color depth of the pixels but also related to the different areas observed by the biological retina.
% The pixels in the center of the field of view are more likely to excite biological nerves, while the pixels in the peripheral area are more difficult.
{To this end,  we propose a novel encoding method for SNNs called Neuronal Receptive Field Encoding (NRFE), which also extracts features from the spatial dimension to achieve high-precision traffic sign recognition.}
More specifically, the main contributions of this paper are summarized as follows.

\begin{itemize}
	\item We propose a new federated learning framework with spike neural networks for efficient and distributed traffic sign recognition. The spike neural networks are particularly suitable and practical for federated learning in IoV because of the advantages of low power consumption, fast inference, and event-driven information processing. 
	
	\item {We propose a new encoding scheme for SNNs based on the neuronal receptive field to extract features from pixels and different regions of traffic signs. Meanwhile, we design a Gaussian receptive field matrix to simulate the area detected by biological neurons.}
	
 	\item {We perform extensive experiments to show the performance of adaptability, robustness, and energy efficiency of the federated SNN algorithm on a real traffic sign recognition dataset by comparing it with baseline schemes. Numerical results demonstrate the advantages of the proposed algorithm regarding noise resistance and energy efficiency.}
\end{itemize}

The rest of this paper is organized as follows. Section \ref{sec-2} introduces related work and Section \ref{sec-3} presents the methodology of FL and SNN. Section \ref{sec-4} describes the proposed federated SNN framework for traffic sign recognition. Numerical results are presented in Section \ref{sec-5} and conclusions are drawn in Section \ref{sec-6}.

\section{related work}\label{sec-2}
\subsection{Autonomous Driving Approaches for Internet of Vehicles}
With advanced embedded devices and communication technologies, more and more intelligent vehicle applications have emerged, such as automatic driving, emergency collision avoidance, and target tracking.
In particular, vehicle automatic driving applications can display warning signs on the dashboard, actively sense information about the surrounding environment outside the driver’s field of vision and, effectively provide target detection and collision avoidance assistance.
For instance, autonomous driving applications require vehicles to quickly and accurately recognize traffic signs on the roadside.
\textcolor{black}{The authors in \cite{4220659} proposed a road sign intelligence system based on Support Vector Machines (SVMs) for detecting and recognizing road signs.
The authors in \cite{8709983} adopted a Region-Convolutional Neural Network (R-CNN) approach for large-scale detection and recognition of traffic signs through automatic end-to-end learning.
The authors in \cite{yang2018deep} added an attention network to the Fast R-CNN model to extract latent features of traffic signs.}
The authors in \cite{arcos2018deep} proposed a traffic sign recognition system based on a space transformer, and quantitatively compared different adaptive and non-adaptive gradient descent methods.

Although these methods improve the perception and navigation strategies of autonomous driving in smart cars, they still face two severe challenges: data privacy and limited computing resources.

\subsection{Federated Learning for Internet of Vehicles}
In IoV, vehicles normally generate a large amount of data through advanced sensors and vehicular applications containing abundant user privacy information, e.g., location information, user preference, and driving trajectory \cite{7042791}. With the help of powerful Machine Learning (ML) methods, these data are analyzed and utilized to achieve intelligent transportation systems \cite{8926369}. Traditional data analysis using centralized ML approaches has high risks of user privacy leakage \cite{8994206}.  As a distributed privacy-preserving machine learning approach, FL has attracted increasing attention and become an important method to achieve multi-vehicle communication with privacy protection in IoV scenarios. Researchers have introduced federated learning into autonomous driving, in which vehicles locally train models and an RSU aggregate trained model parameters for AI-based decisions in IoV. Such as the authors in \cite{liu2020privacy} designed a federated learning-based gated recurrent unit neural network algorithm for efficient traffic flow prediction in IoV. 
\textcolor{black}{The authors in \cite{8843942} proposed a Differential Privacy-based Asynchronous Federated Learning (DPAFL) scheme for resource sharing and secure local model updates in IoV.
The authors in \cite{9378811} designed an FL-based vehicle cooperative positioning system to achieve high-precision coordinated vehicle positioning while ensuring user privacy.}
To ensure secure and reliable FL, Kang \textit{et al.} in \cite{8832210} utilized blockchain to achieve secure reputation management, and then proposed a reputation-based worker selection scheme to encourage high-quality workers to participate in FL. Furthermore, Lu \textit{et al.} in \cite{lu2020blockchain} designed a blockchain empowered FL architecture to achieve secure data sharing without heavy transmission load and privacy risks. 
Hu \textit{et al.} in \cite{hu2021towards} build a two-layer federated structure by coordinating multiple RSUs, which reduces the computational overhead of traffic flow and protects data privacy during model training.

Due to the limited computing resource and energy constraints for vehicular networks, the above FL methods do not work well in resource-hungry and complex vehicular applications.

\begin{figure}[!t]
	\centering
	\includegraphics[width=1\linewidth]{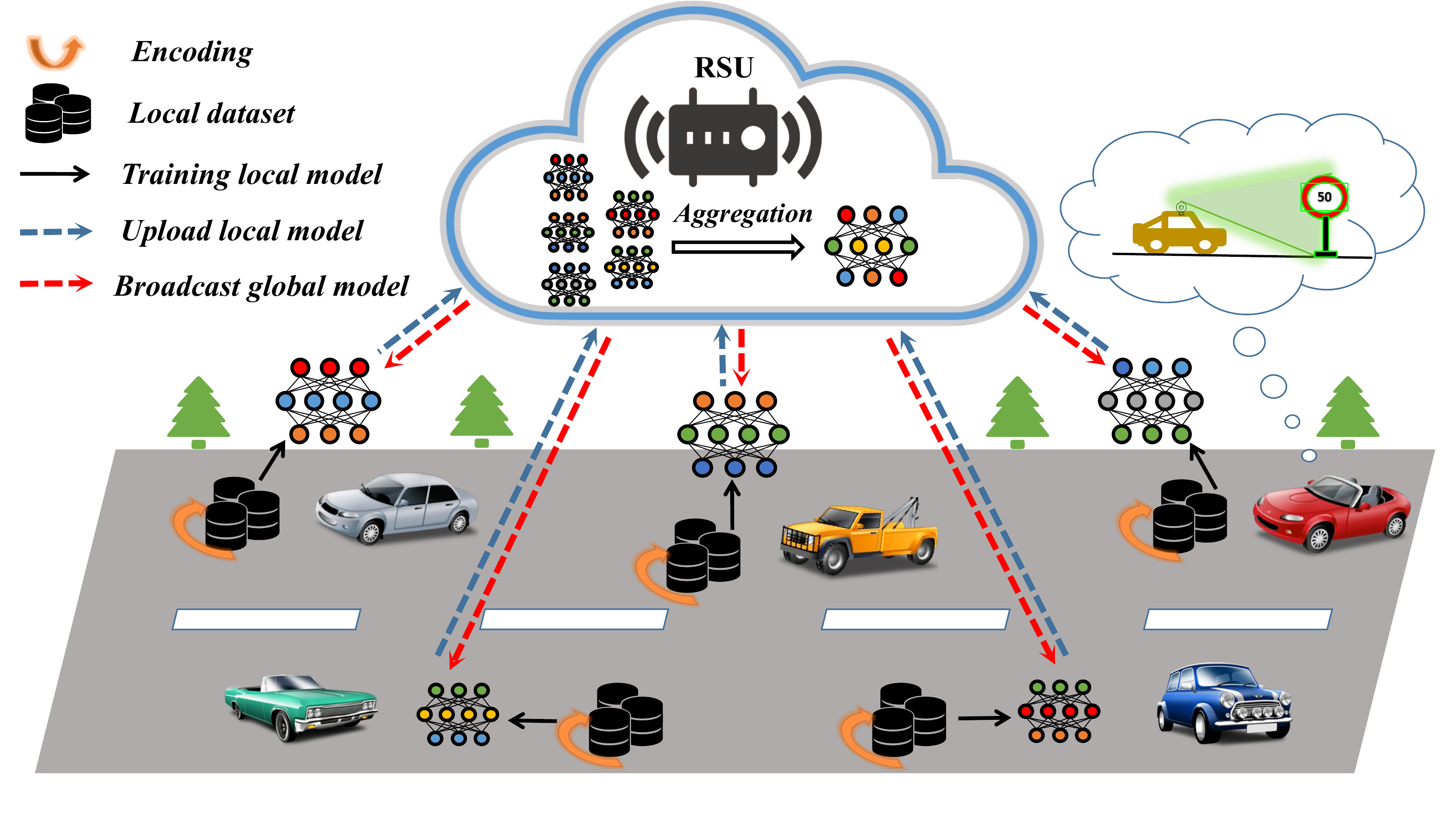}
	\caption{\textcolor{black}{Overview of federated spike neural networks in autonomous driving for road traffic sign recognition.}}
	\label{Model}
\end{figure}

\subsection{Low-power Spike Neural Network}
Deep Convolutional Neural Networks (CNNs) provide efficient tools for the accurate and effective processing of vehicular data. 
However, it requires a tremendous amount of computing resources,  time, and energy for task training, which may be not practical and feasible for IoV scenarios with limited computing resources and strict energy consumption restrictions \cite{9053856,kheradpisheh2018stdp}. To address these challenges, spike neural networks, which are next-generation artificial neural networks inspired by information processing in biology, have been introduced to efficiently achieve deep neural network implementation with favorable properties including low-power consumption, fast inference, and event-driven information processing \cite{pfeiffer2018deep}. Chandarana \textit{et al.} in \cite{9651610} proposed a method of encoding a static image into a time pulse sequence and achieving adaptive signal sampling for low-power edge computing through SNNs.
Kim \textit{et al.} in \cite{kim2020revisiting} proposed a time batch normalization technique to train high-precision and low-latency SNN, which brings huge energy efficiency benefits to the hardware. Therefore, SNNs are becoming a promising alternative to traditional convolutional neural networks in IoV. 
For example, the author in \cite{bing2020indirect} proposed Reward-modulated Spike-Timing-Dependent Plasticity (R-STDP), which combines reinforcement learning and spike-timing-dependent plasticity in IoV for end-to-end training SNN.
The author in \cite{7862386} proposed a spike neural network based on visual coding inspired by silicon retina, which controls the end-to-end lane following vehicle behaviors.

Inspired by previous work, this paper studies the traffic sign recognition tasks, which require strong privacy protection, and considers vehicles with limited computing resources and energy in IoV. To this end, we propose a federated spike neural network framework, which provides low power consumption and high-precision traffic sign recognition while ensuring user privacy. \textcolor{black}{Furthermore, the symbols of this paper are summarized in Tab. \ref{symbols}.}

% \vspace{-0.2cm}
\begin{table}[!t]
\scriptsize
	\centering
	\color{black}\caption{Summaries of applied symbols.}
	\begin{tabular}{|c|c|}\hline
		\textcolor{black}{Symbols} & \textcolor{black}{Description}\\\hline
        $\mathcal{K}_r$ & The set of clients selected in round $r$\\
		$\mathcal{D}_k$ & The $k$-th client local dataset\\
		$\mathcal{O}_{k}$ & The dataset of the $k$-th client after encoding\\
		$\mathcal{L}_k$ & The $k$-th client local loss function\\
		$L(\cdot)$ & The cross-entropy function\\
		$\omega_{r}$ & The global model of round $r$\\
		$\omega_{r}^{k}$ & The local model of the $k$ client in round $r$\\
		$\omega^\ast$ & The global optimal model\\
	    $f_\omega ( \cdot )$ & The prediction function on model $\omega$\\
		$x_i$ & The $i$-th sample\\
		$o_i$ & The $i$-th sample to be encoded\\
		$y_i$ & The label of the $i$-th sample\\
		$p_k$ & The contribution rate of the $k$-th client\\
		$c$ & The total number of classes\\
		$K$ & The total number of clients\\
		$F$ & The client participation rate\\
		$B$ & The local min-batch size\\
		$E$ & The number of local epochs\\
		$M$ & The side length of the output feature map\\
		$N_l$ & The number of neurons in layer $l$\\
		$E(l)$ & The energy consumption of layer $l$\\
		$r$ & The round of communication\\
		$ks$ & The convolution kernel size\\
		$s$ & The input feature size\\
        $C$ & The number of channels\\
        $I(t)$ & The input current at time $t$\\
        $u(t)$ & The membrane potential at time $t$\\
        $u_{rest}$ & The resting potential\\
        $u_{i}^{t}$ & The membrane potential of the $i$-th neuron at time step $t$\\
        $o_{i}^{t}$ & The spike sequence value of neuron $i$ at time step $t$\\
        $\gamma_{i}^{t}$ & The learned weight of the BN layer\\
		$\theta$ & The threshold voltage of spike neural network\\
		$t$ & The time step of spike neural network\\
		$\lambda$ & The leakage factor\\
		$\alpha$ & The substitution rate\\
		$w_{ij}$ & The weight of the connection between neuron $i$ and neuron $j$\\
		$W_l$ & The weight of the $l$-th layer\\
		$R_{s} (l)$ & The spike rate of the $l$-th layer\\
		$\tau_{m}$ & The time constant of membrane potential decay\\
		$\mu$ & The parameters of Dirichlet distribution function\\\hline
	\end{tabular}
	\label{symbols}
\end{table}

\section{methodology}\label{sec-3}
\subsection{Federated Learning}
Federated Learning (FL) is a distributed machine learning framework with privacy protection, which requires the transmission of model parameters rather than raw data between clients and the server \cite{liu2020privacy}.
In FL, we assume that there are $K$ clients and a server $\mathcal{S}$,  which cooperate to train a global optimal model $\omega^\ast$, where the server does not hold any raw data while each client holds a local dataset $\mathcal{D}_k$.
The detailed training process of FL is described in the following.

\textcolor{black}{In each communication round $r$, the server randomly selects a subset of clients $\mathcal{K}_r \subseteq \mathcal{K}$ to send the current global model $\omega_r$.
Then, the selected clients use Stochastic Gradient Descent (SGD) to train the received global model $\omega_r$ on their local dataset $\mathcal{D}_k = \{(x_1, y_1), (x_2, y_2), \cdots, (x_i, y_i)\}$.
For the $k$-th client, the following objective function needs to be minimized:
\begin{equation}
	{\mathcal{L}_k}(\omega) = \dfrac{1}{|\mathcal{D}_k|} \sum\nolimits_{{x_i},{y_i} \in {{\cal D}_k}} L_{k} (y_{i}, f_{\omega_{k}}(x_{i})),
\end{equation}
where $L_{k}(\cdot)$ generally is the cross-entropy function, $y_{i}$ is the ground-truth of $x_{i}$, $f_{\omega_{k}}(\cdot)$ is the probability vector predicted by the model. 
Meanwhile, the local update of the $k$-th client is as follows:
\begin{equation}
	\omega_{r+1}^{k} = \omega_{r} - \eta\bigtriangledown \mathcal{L}_{k}(\mathcal{D}_k; \omega),
\end{equation}
where $\eta$ is learning rete. After completing local training, each client uploads its local model update $\omega_{r+1}^{k}$ to the server.}

\textcolor{black}{Next, all uploaded model updates will be aggregated into the current global model by the server using the federated average algorithm (i.e., FedAvg \cite{mcmahan2017communication}), which is expressed as:
\begin{equation}\label{eq-2}
    \omega_{r+1} = \sum\nolimits_{k \in {\mathcal{K}_r}} {p_k} \omega_{r+1}^{k},
\end{equation}
where $p_k=\frac{|D_k|}{\sum\nolimits_{k\in {\mathcal{K}_r}}|D_k|}$ represents the contribution rate of the $k$-th client to the current global model, $\mathcal{K}_r$ represents the client set selected in the $r$-th round.}

\textcolor{black}{In the whole FL training process, with the increase of rounds, the model will tend to converge to obtain the optimal global model $\omega^\ast$.}
\subsection{Spiking Neural Networks}
\begin{figure}[t] 	\centering 	\subfigure[]{\includegraphics[width=0.45\linewidth]{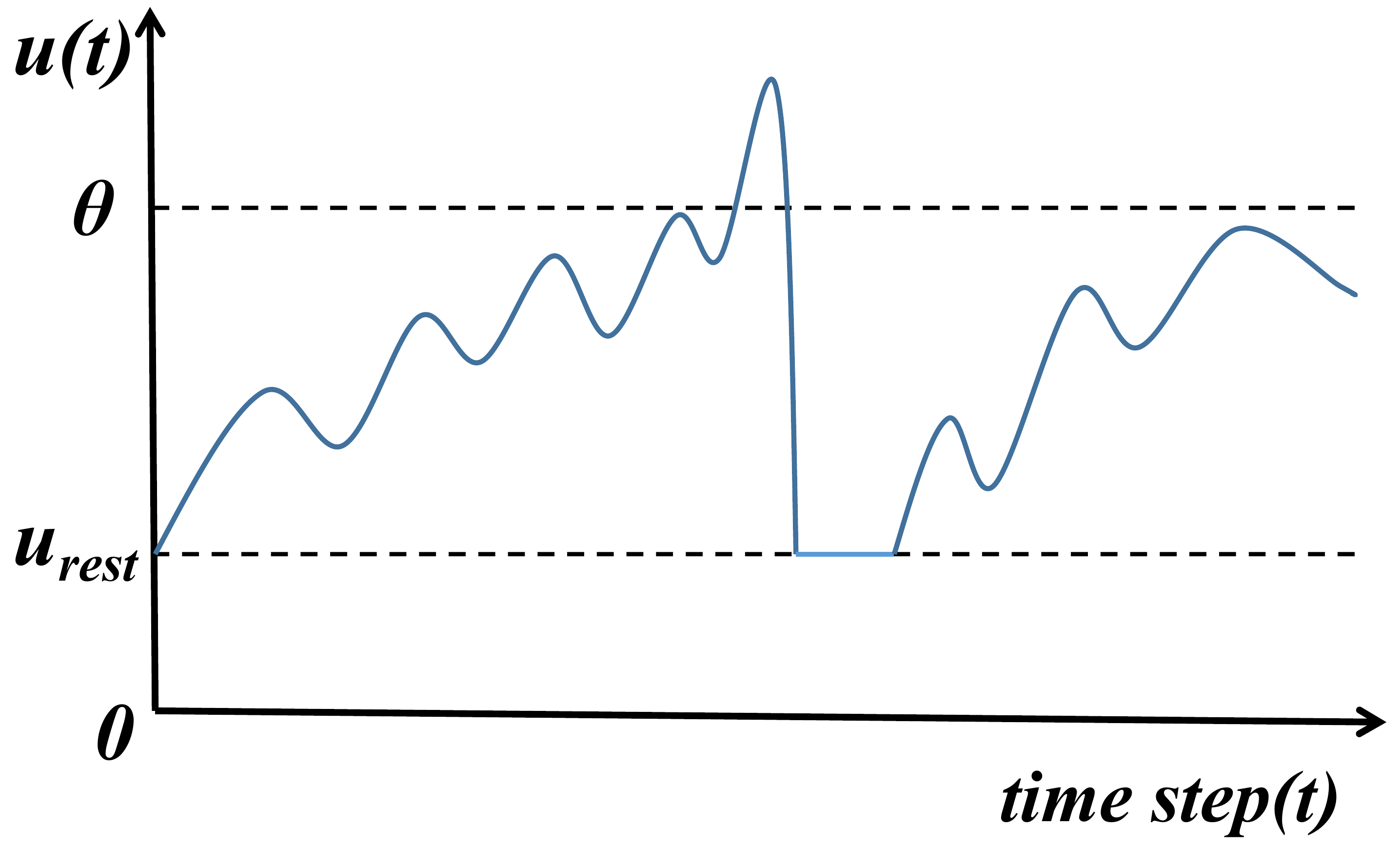} 		\label{a-11}} 	\subfigure[]{\includegraphics[width=0.45\linewidth]{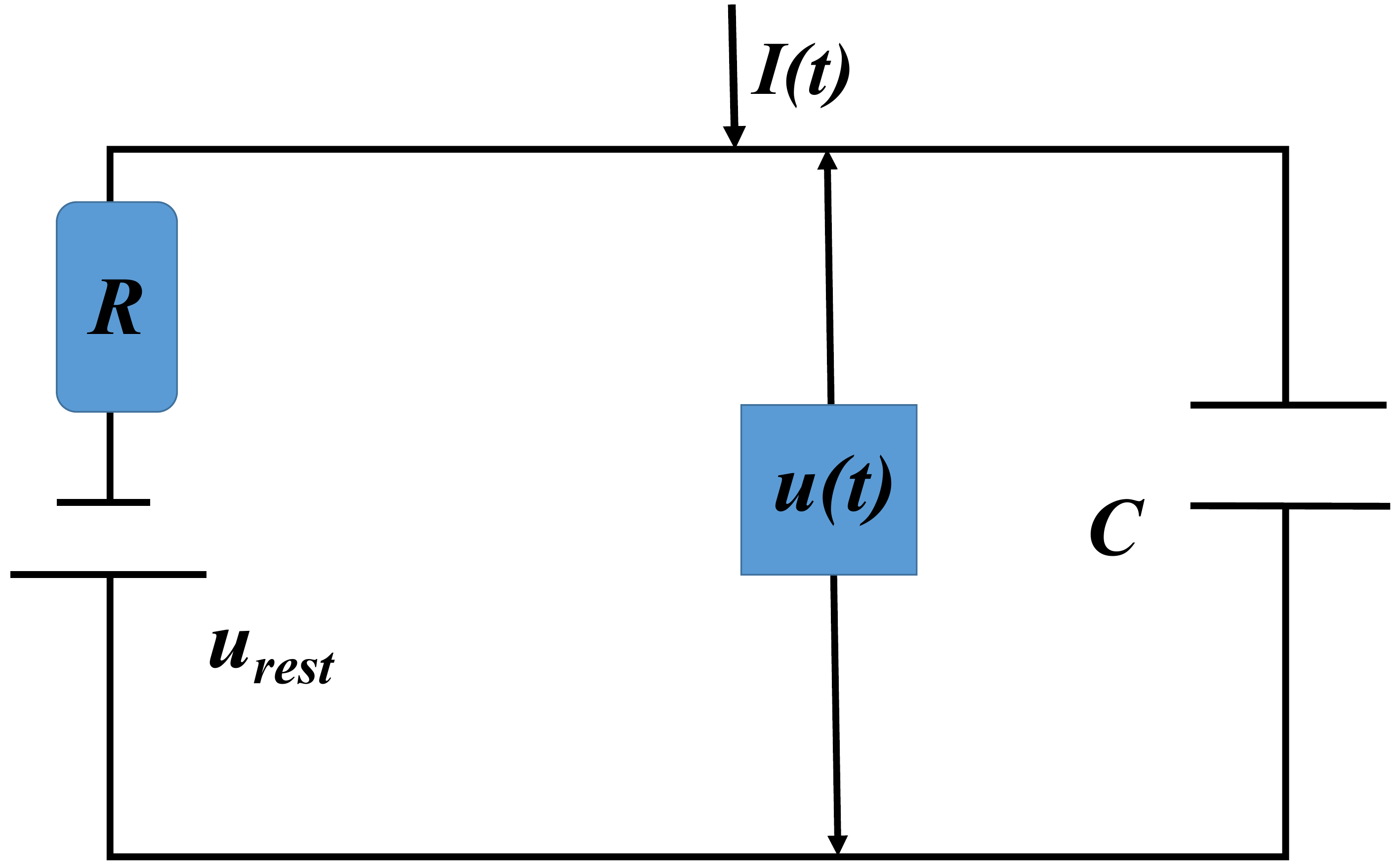} 		\label{b-12}} 	\caption{(a) Changes in membrane potential; (b) Physical structure of LIF model.} 	\label{fig} \end{figure}

{Spiking Neural Networks (SNNs) have emerged as a promising computing paradigm using discrete action potentials (i.e., spikes) that highly mimic the physiological mechanisms by which biological neurons propagate information \cite{roy2019towards}.
Unlike traditional Artificial Neural Networks (ANNs) \cite{7886285,7015548,9384272}, SNNs use binary spikes to transmit information between layers and layers of the network with less computational cost and extremely low power consumption \cite{bouvier2019spiking}.}
\textcolor{black}{The low power consumption of SNNs is reflected in the fact that the floating-point product operation in ANNs can be simplified to the accumulation operation of multiple additions \cite{9583900}.}
Specifically, each neuron maintains a membrane potential level, which integrates the input spike signal \cite{xiao2021training}. 
When a neuron receives a sequence of spike signals within a time step $t$, its membrane potential will increase, decrease, or remain unchanged.
Once the membrane potential of the neuron exceeds a threshold $\theta$, it generates an output spike signal $1$ to the next layer of neurons. 
Next, the neuron enters a transient refractory period. 
At this time, no matter whether any spike information is received, the membrane potential will not change.
If the neuron does not reach the threshold potential within the time step, it will output a spike signal $0$.
In addition, the membrane potential will also leak during the accumulation process.
Fig. \ref{a-11} shows the change of membrane potential of a spiking neuron in time step $t$.

To simulate the spike generation process of neurons, researchers widely use Leakage-Integration-and-Fire (LIF) model
\cite{dayan2005theoretical}, as shown in Fig. \ref{b-12}, which is similar to a physical structure composed of a capacitor $C$, a resistor $R$, power supply $u$, and an input current $I$.
Here, we regard the input current $I(t)$ as a spike sequence received by neurons, and the process of accumulating neuronal membrane potential as a charging capacitor $C$. 
Namely, when capacitor $C$ accumulates sufficient charge and reaches the threshold voltage $\theta$, it will enter the discharge state, i,e., a spike signal will be generated.
If the input current $I(t)$ disappears, the voltage of the entire capacitor is provided by $u_{rest}$, which is the resting potential.
Moreover, the charge leaks from the cell membrane when neurons accumulate membrane potential, so a leakage resistance $R$ can be used to simulate this biological phenomenon.
\textcolor{black}{According to the law of conservation of current, the input current can be divided into two parts: $I(t) = I_{R}(t) + I_{C}(t)$.}
The LIF can be expressed as a differential equation:
\begin{equation}
\tau_{m} \frac{du(t)}{dt} = -(u(t)-u_{rest})+RI(t), u(t)< \theta,
\end{equation}
where $\tau_{m} = RC$ is the time constant of membrane potential decay,
$u(t)$ represents the neuron membrane potential at time $t$, $u_{rest}$ is the resting potential, $\theta$ is the threshold voltage. Also, $R$ and $I(t)$ represent the resistance and input current at time $t$, respectively.

\begin{figure}[!t]
	\centering
	\includegraphics[width=0.8\linewidth]{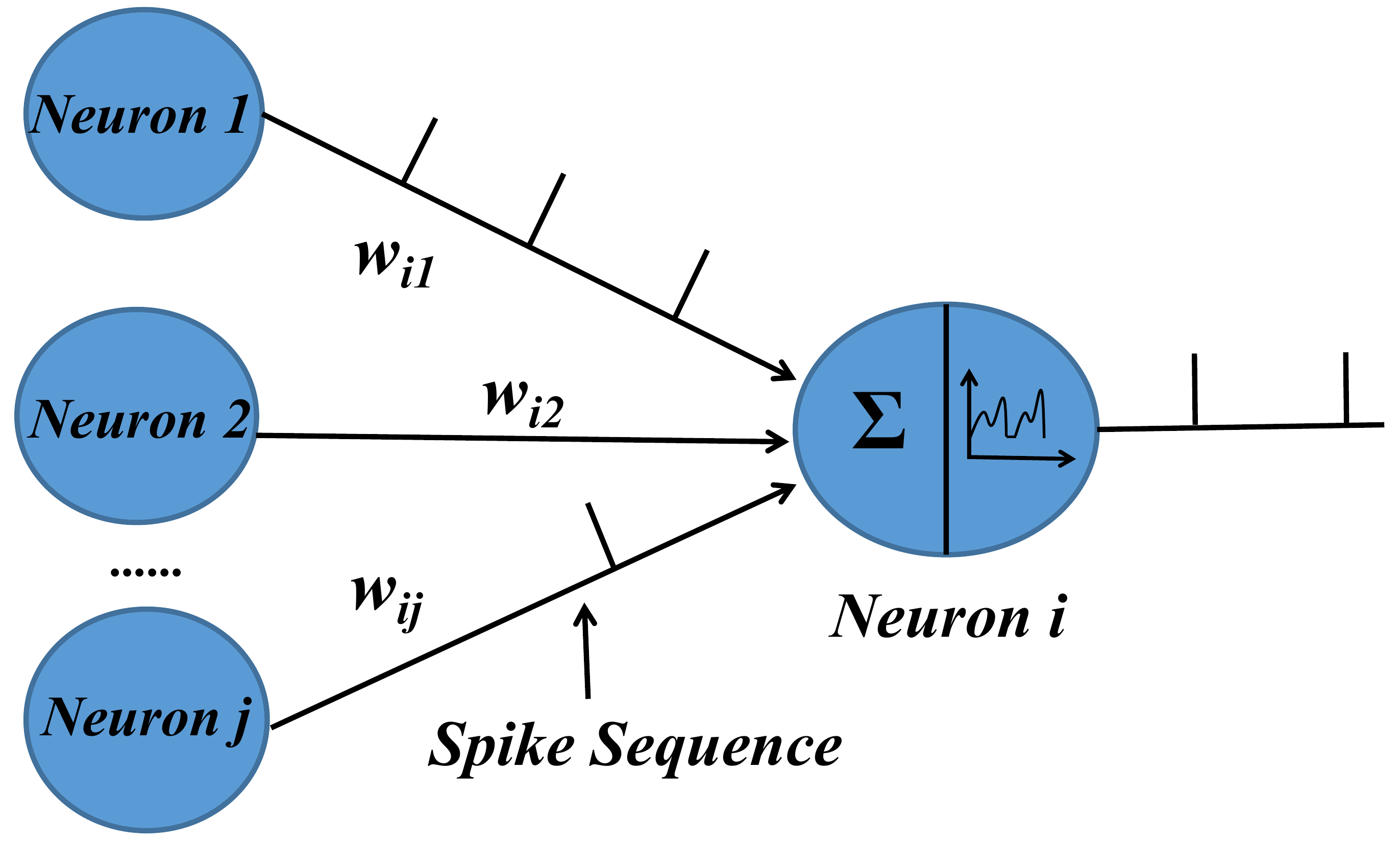}
	\caption{Spike neuron propagation process.}
	\label{Neuron}
\end{figure}

As illustrated in Fig. \ref{Neuron}, specific to a single neuron $i$, we can express the membrane potential $u_{i}^{t}$ at time step $t$ as: 
\begin{equation}
u_{i}^{t} = \lambda u_{i}^{t-1} + \sum\limits_{j} w_{i j} o_{j}^{t},
\end{equation}
where $\lambda \in (0, 1)$ is the leakage factor, $j$ represents the number of neurons in the previous layer, $w_{i j}$ represents the weight of neuron $i$'s connection to neuron $j$ in the previous layer. Here, $o_{j}^{t}$ indicates that neuron $i$ receives the spike sequence of neuron $j$ at time $t$. At this point, the spike sequence output by neuron $i$ at time $t$ can be expressed as:
\begin{equation}
o_{i}^{t} = \begin{cases}1, &{if}\ u_{i}^{t}\ge\theta\\0,&{otherwise}\end{cases}.
\end{equation}
Note that the spike neural network has no bias $b$ in the forward propagation since it adds redundant voltage to the membrane potential of SNNs \cite{kim2020revisiting}.

For the output layer, we discard the function of neurons triggering spikes so that neurons do not generate any spikes.
Specifically, when the spike signal propagates to the last layer of the SNN, the membrane potential in the accumulation time step $t$ will be output.
Similar to ANNs, the number of neurons in the last layer of SNNs is also determined by the total number of classes in the training dataset.
In this way, the output layer can predict the probability distribution of samples without causing information loss.
Thus, the cross-entropy loss function of SNNs can be expressed as:
\begin{equation}\label{eq-6}
L = -\sum_{j=1}^{c} y_{j}log(\frac{e^{u_{j}^{t}}}{\sum_{j=1}^{c}e^{u_{j}^{t}}}),
\end{equation}
where $y$ is the ground-truth label, $c$ is the total number of classes in the training dataset, {$u_{j}^{t}$ is the membrane potential accumulated by the $j$-th neuron in the output layer of the SNN during a time step $t$.}

For model updates, SNNs usually use gradient descent to backpropagate the loss value to update the weights of all layers.
Unlike ANNs, SNNs need to calculate the gradient of a time step.
According to the chain derivation rule, the gradient of the $l$-th layer can be defined as:
\begin{equation}\label{eq-7}
\bigtriangleup W_{l} = \begin{cases}\sum\limits_{t=1}^{t} \frac{\partial L}{\partial O_{l}^{t}} \frac{\partial O_{l}^{t}}{\partial U_{l}^{t}} \frac{\partial U_{l}^{t}}{\partial W_{l}^{t}}, &{if}\ l = hidden\ layer\\\sum\limits_{t=1}^{t} \frac{\partial L}{\partial U_{l}^{t}} \frac{\partial U_{l}^{t}}{\partial W_{l}^{t}}, &{otherwise}\end{cases},
\end{equation}
where $U_{l}$ is the membrane potential of neurons in the $l$-th layer of the SNN, $O_{l}$ is the spike sequence of the $l$-th layer, and $W_{l}$ is the weight of the $l$-th layer.
However, in SNNs, neurons can only produce spike output when the membrane potential exceeds the threshold, resulting in non-differentiable.
Thus, to overcome this, researchers generally use some alternative functions to approximate the derivative of the threshold function, the most common of which is the piece-wise linear function, such as:
\begin{equation}
\frac{\partial o_{i}^{t}}{\partial u_{i}^{t}} = \alpha \max \{0, 1-|\frac{u_{i}^{t} - \theta}{\theta}|\},
\end{equation}
where $\alpha$ is the substitution rate that controls the update amplitude of SNNs, $\theta$ is the threshold voltage.
The reason for this design is that when the membrane potential of a neuron is closer to the voltage threshold, the membrane potential is easier to be activated and send spikes to the next layer of neurons.
On the other hand, since the gradient is accumulating at each time step, we guess that the value of the hyperparameter $\alpha$ is also related to the time step $t$.
For a larger time step $t$, a smaller hyperparameter $\alpha$ should be set to avoid gradient explosion.
% Through experimental verification, we observed that for a larger time step $t$, a smaller hyperparameter $\alpha$ should be set.
Overall, the backpropagation method of SNNs is similar to ANN, except that the alternative function is used to approximate the non-differentiable threshold function.
Therefore, the model update of the $l$-th layer of SNNs can be expressed as: $W_{l} = W_{l} - \eta\bigtriangleup W_{l}$.

\subsection{Training Spiking Neural Networks with Batch Normalization Through Time}
Batch Normalization (BN) \cite{santurkar2018does} has the advantages of accelerating the speed of network training and convergence, controlling gradient explosion, and preventing gradient disappearance, which has been widely used in the deep learning model.
However, the BN layer only normalizes a batch of inputs, ignoring the information in the time dimension.
Thus, the naive application of the BN layer in SNN cannot capture the time representation of the spike sequence.
Recently, the authors in \cite{kim2020revisiting} propose a method to update the internal parameters of the BN layer with the time step to improve the SNN training performance, which is called Batch Normalization Through Time (BNTT).
Specifically, after the spike sequence passes through the convolution or linear layer, it will enter the BN layer that expands with the time step $t$, i.e., $t$ BN layers.
The mathematical formula of BNTT is given as follows:
\begin{equation}
\begin{aligned}
u_{i}^{t} = \lambda u_{i}^{t-1} + BNTT_{\gamma^{t}}\sum\limits_{j} w_{i j} o_{j}^{t},\\
=\lambda u_{i}^{t-1} + \gamma_{i}^{t} (\frac{\sum_{j} w_{i j} o_{j}^{t} - \mu_{i}^{t}}{\sqrt{(\sigma_{i}^{t})^{2}+\epsilon}}),
\end{aligned}
\end{equation}
where $\gamma_{i}^{t}$ is the learned weight of the BN layer, $\epsilon$ is a constant that prevents the denominator from being $0$, $\mu_{i}^{t}$ and $\sigma_{i}^{t}$ are the mean and variance from the samples in a mini-batch $B$ for time step $t$, respectively.
For the update of the BN layer, given $x_{i}^{t} = \sum\limits_{j} w_{i j} o_{j}^{t}$, the gradient of weight $\gamma^{t}$ in a batch $b$ can be expressed as:
\begin{equation}\label{eq-10}
\bigtriangleup \gamma^{t} = \frac{\partial L}{\partial u^{t}} \frac{\partial u^{t}}{\partial \gamma^{t}} = \sum\limits_{i=1}^{b} \frac{\partial L}{\partial u^{t}} \hat{x}_{i}^{t}.
\end{equation}
{Here, $\hat{x}_{i}^{t} = \frac{x_{i}^{t} - \mu^{t}}{\sqrt{(\sigma^{t})^{2}+\epsilon}}$. Therefore, a batch of weight updates for the $t$-th BN layer can be calculated as:}
\begin{equation}
  \gamma^{t} = \gamma^{t} - \eta\bigtriangleup \gamma^{t}.  
\end{equation}

\section{Federated spike neural network encoded by neuronal receptive field}\label{sec-4}
\subsection{Neuronal Receptive Field Encoding}
{The attention mechanism originates from the fact that humans can reasonably utilize limited visual information processing resources to selectively focus on visual focus areas, and has been widely used in the field of deep learning.
In recent years, many attention models such as Transformer \cite{jaderberg2015spatial}, DIN \cite{zhou2018deep} and AFM \cite{xiao2017attentional} have been proposed in natural language processing and image recognition.
Inspired by the attention mechanism, we believe that information in the central region of the neuron receptive field is more likely to activate that neuron to emit spikes, while it is more difficult for those at the edges.
To this end, we propose a novel encoding method for SNNs called Neuronal Receptive Field Encoding (NRFE) for fast and efficient training.
Specifically, we design a Gaussian distribution matrix to simulate the regions observed by neurons, and then combine the pixel features of the image to jointly encode the local traffic sign dataset of vehicles.
Next, we provide more details about the NRFE method as follows.}

\begin{itemize}
    \item \textbf{\textit{Step 1, Normalization:}} The input feature of an image is usually composed of three channels of RGB pixels.
    Note that the range of RGB pixel values is between $0$ and $255$.
    However, if the input features are large, the gradient passed to the input layer during backpropagation will become large, resulting in a suboptimal model.
    Therefore, in order to accelerate the model convergence and prevent gradient explosion, we normalize the input features.
    For each channel's input $X = \begin{bmatrix}x_{ij}\end{bmatrix}_{s \times s}$, we normalize it to the range $0$ to $1$ (i.e., $X \to X^ \ast$).
    \begin{equation}
    x_{ij} ^\ast = \frac{x_{ij}-x_{min}}{x_{max}-x_{min}},
    \end{equation}
    where $x_{max}$ and $x_{min}$ are the minimum and maximum pixel values in matrix $X$, respectively.
    Here, $s$ is the size of the input image.
    \item \textbf{\textit{Step 2, Gaussian receptive field matrix:}} The receptive field is the output feature of each layer in SNNs, which is mapped to the size of the area on the input image, i.e., a point on the output feature corresponds to an area on the input image.
    In the receptive field, the pixel is closer to the center, and the greater contribution to the calculation of output features will be.
    In other words, the closer the pixel is to the center of the receptive field, the more times it will be convolved.
    However, the contribution of all pixels in the receptive field to the output feature is not all the same. In many cases, the influence of pixels in the receptive field area on the model follows a Gaussian distribution \cite{luo2016understanding}.
    Inspired by \cite{bohte2002error}, we design a Gaussian receptive field matrix $G_{s \times s} \sim N(\mu_{g}, \sigma_{g}^{2})$ to simulate the area detected by biological neurons, as follows.
    \begin{equation}
    \mu_{g} = \frac{1}{s} \sum\limits_{i=1}^{s} \frac{2i-1}{2(s-ks)},
    \end{equation}
    \textcolor{black}{\begin{equation}
    \sigma_{g} = \frac{BCE}{s-ks},
    \end{equation}}
    where $ks$ represents the size of the convolution kernel, $C$ represents the number of input channels, $B$ is the local mini-batch size, and $E$ is the local epoch.
    \item \textbf{\textit{Step 3, Coding:}} 
    For an image, we consider that pixels in the center of the receptive field and with large values are more likely to activate neurons to emit spike signals.
    Thus, we encode such pixels as $1$.
    Secondly, for the pixels with small values at the edge of the receptive field, we encode them as $-1$.
    In other cases, it is coded as $0$.
    Given $Q = X^\ast + G$, the coding formula of the Gaussian receptive field is as follows:
    \begin{equation}
    input =\begin{cases}1, &q_{ij} \geqslant x_{ij} ^\ast \cap q_{ij} \geqslant g_{ij}
    \\-1, & q_{ij} \le x_{ij} ^\ast \cap q_{ij} \le g_{ij}
    \\0, & otherwise\end{cases},
    \end{equation}
    where $q_{ij} \in Q$, $x_{ij} ^\ast \in X^ \ast$, $g_{ij} \in G$.
\end{itemize}

\subsubsection{Discussion}
{Different coding schemes have different principles for extracting features from the raw data.
The choice of a specific encoding method depends on whether the data has relevant information in the temporal and spatial domains, and whether there is noise.
In rate-based coding of SNNs \cite{wu2019direct}, the input of each image is coded as a sequence of Poisson spikes according to its pixel intensity.
The probability of a spike in this sequence is positively correlated with the pixel intensity of the image.
% In other words, the greater intensity of a single pixel of an image leads to more spikes emitted by its encoded spike sequence in the time step $t$.
However, it generally requires very high pixel values to be encoded into a sequence of high-frequency spikes for quick propagation forward in the network.
Therefore, rate coding has limitations in the face of noisy and low-pixel traffic sign data.
Another common encoding method is temporal-based \cite{9416238}, which generates the precise time of each spike according to changes in the input signal.
Although this approach has been shown to be effective in practice, it is generally suitable for the fast processing of time series and streaming data.
Thus, it is also inappropriate for traffic sign recognition.
Compared with the above two encoding methods, our proposed encoding scheme based on neuron receptive fields extracts features from pixel and spatial dimensions and is especially suitable for encoding traffic sign data.}

\begin{algorithm}[!t]\label{algorithm 1}
	\color{black}\caption{\textcolor{black}{$\mathrm{FedSNN}$-$\mathrm{NRFE}$ algorithm.}}
	\begin{algorithmic}[1]
	\REQUIRE Set of $\mathcal{K}$ clients, mini-batch size $B$, local training epoch number $E$, time step $T$, network layer $L$, and learning rate $\eta$.
	\ENSURE The convergent SNN model $\omega^\ast$.
	\STATE \textbf{Server executes:}
	\STATE Initialize a global SNN model $\omega_0$
	\FOR{each round $r = 0, 1, 2, ...$}
% 		\STATE $m \gets \mathrm{max}(F\cdot K, 1)$\\
		\STATE $S_r \gets$ randomly select $m=\mathrm{max}(F\cdot K, 1)$ clients from a set of $\mathcal{K}$ clients\\
		  \FOR{each client $k \in S_r$ $\textbf{in parallel}$}
		  \STATE $\omega_{r+1}^{k} \gets$ \textbf{ClientUpdate($k, \omega_{r}$)}\\
		  \ENDFOR
		\STATE$\omega_{r+1} = \sum\nolimits_{k \in {\mathcal{S}_r}} {p_k} \omega_{r+1}^{k}$\\
	\ENDFOR
		\STATE \textbf{ClientUpdate($k$, $\omega_{r}$):}$//$ For the $k$-th client\\
		\STATE$\mathcal{O}_{k}$ $\gets$ NRFE $(\mathcal{D}_k)$
		\FOR{each local epoch $e \gets 1$ to $E$}
		    \FOR{mini-batch  $B$}
		        \FOR{layer $l$ from $1$ to $L-1$}
		            \FOR{time step $t$ from $1$ to $T$}
		                \STATE$U_{l}^{t} = \lambda U_{l}^{t-1} + BNTT_{\gamma^{t}} (W_{l}, O_{l-1}^{t-1})$
		            \ENDFOR
		        \ENDFOR
		        \STATE $//$Accumulation of membrane potential at layer $L$
		        \STATE $\omega_{r+1}^{k} = \omega_{r} - \eta\bigtriangledown \mathcal{L}_{k}(k, \mathcal{O}_{k}; \omega_{r})$
		    \ENDFOR
		\ENDFOR
		\RETURN $\omega^\ast$ to server.
	\end{algorithmic}
\end{algorithm}

\subsection{Federated Spike Neural Networks for IoV}
Autonomous driving applications in IoV may involve multiple fields and need to integrate data from various agencies and departments.
However, the data collected by vehicles usually include private information such as driver location and traveling trajectory.
Therefore, it is generally unrealistic to collect all data on a centralized server.
Federated learning is essentially an encrypted distributed machine learning technology, which provides an effective computing paradigm for executing multi-vehicle autonomous driving tasks.
Furthermore, due to the limited computing and storage resources of most IoV devices (e.g., vehicles), it is impossible to achieve complex artificial intelligence computing tasks.
In contrast, SNNs achieve comparable performance to deep neural networks with relatively low computational cost and extreme energy efficiency, which has been widely used in IoV, such as image classification, target detection, navigation, and motion control. 
Thus, based on the automatic driving scenarios of IoV, we propose a federated spike neural network algorithm called $\mathrm{FedSNN}$-$\mathrm{NRFE}$, which focuses on a fast and effective coding scheme to achieve low-power and high-performance FL.

Considering that a group of wirelessly connected vehicles perform FL autonomous driving tasks on the same road, as shown in Fig.~\ref{Model}.
These vehicles use on-board cameras to scan traffic signs in front of the road and then make appropriate driving decisions.
The entire FL-base IoV system consists of a roadside unit (RSU, i.e., server) and $K$ vehicles working as clients, in which each  vehicle has its own private local dataset $\mathcal{D}_k$.
% The whole learning process is similar to that of  traditional FL, except that the local vehicle is trained by the SNN model.
The RSU aggregates the SNN models uploaded by the  vehicle and sends the global model. 
Each vehicle uses the local dataset to train the SNN model.
Specifically, for communication round $r$, the iterative process of $\mathrm{FedSNN}$-$\mathrm{NRFE}$ is based on the following protocol:
\begin{itemize}
  \item [1)] 
   The RSU randomly selects a certain proportion of $F$ ($0<F<1$) vehicles from all vehicles to deliver the global model parameter $\omega_r$. 
  \item [2)]
   \textcolor{black}{For the $k$-th  vehicle, it first utilizes the NRFE method to encode a local private dataset, i.e.,
   \begin{equation}
       NRFE(\mathcal{D}_k) \to \mathcal{O}_{k},
   \end{equation}
   where $\mathcal{O}_{k} = \{(o_1, y_1), (o_2, y_2), \cdots, (o_i, y_i)\}$.
   Afterwards, the encoded dataset $\mathcal{O}_{k}$ is used to train the received global SNN model $\omega_r$. The loss function is defined as follows:.
   \begin{equation}
       {\mathcal{L}_k}(\omega) = \dfrac{1}{|\mathcal{O}_{k}|} \sum\nolimits_{{o_i},{y_i} \in {{\mathcal O}_k}} L_{k} (y_{i}, f_{\omega_{k}} (o_{i})),
   \end{equation}
   where $L_{k}(\cdot)$ is the cross-entropy function in Equation \eqref{eq-6}, $f_{\omega_{k}} (o_{i})$ is the $softmax$ output of the accumulated membrane potential in the last layer of the SNN model, i.e., $f_{\omega_{k}} (o_{i}) = \{f_{\omega_{k}} (o_{i})_{1}, \cdots, f_{\omega_{k}} (o_{i})_{j}, \cdots, f_{\omega_{k}} (o_{i})_{c}\}$. Specifically, the $softmax$ of $f_{\omega_{k}} (o_{i})_{j}$ is expressed as follows:
   \begin{equation}
       f_{\omega_{k}} (o_{i})_{j} = \frac{e^{u_{j}^{t}}}{\sum_{j=1}^{c}e^{u_{j}^{t}}}.
   \end{equation}
   Given a time step $t = 1, 2, \cdots, T$, the membrane voltage $U_{l}^{t}$ for each layer $l = 1, 2, \cdots, L-1$ can be expressed as:
   \begin{equation}
       U_{l}^{t} = \lambda U_{l}^{t-1} + BNTT_{\gamma_{l}^{t}} (W_{l}, O_{l-1}^{t-1}).
   \end{equation}
   Similarly, when $U_{l}^{t}$ is greater than the threshold voltage $\theta$, $O_{l}^{t} = 1$; otherwise, $O_{l}^{t} = 0$.
   Then, the vehicle computes the gradient change of the local model (Equations \eqref{eq-7} and \eqref{eq-10}) based on the loss to update the model weights.
      \begin{equation}
       \omega_{r+1}^{k} = \omega_{r} - \eta\bigtriangledown \mathcal{L}_{k}(k, \mathcal{O}_{k}; \omega_{r}).
   \end{equation}}
  \item [3)]
   The RSU collects the model parameters uploaded by the local vehicle and uses the federated average aggregation algorithm (Equation \eqref{eq-2}) to update the global model.
\end{itemize}

Note that $\mathrm{FedSNN}$-$\mathrm{NRFE}$ repeats all the above steps until the global model converges or reaches a predetermined accuracy.
The proposed $\mathrm{FedSNN}$-$\mathrm{NRFE}$ algorithm training details are given in Algorithm 1. 
{Furthermore, for the image recognition task of traffic signs, the complexity mainly depends on the convolutional layers of SNNs.
Thus, given the number of convolutional layers $l$, the size of the convolutional kernel $ks$, the side length $M$ of the output feature map, and the number of input and output channels $C_{in}$ and $C_{out}$, the complexity of this algorithm is $O(rFK \sum\limits_{l} ks_{l}^{2} M_{l}^{2} C_{in}^{l} C_{out}^{l}T)$.}

\begin{figure}[!t]
	\centering
	\includegraphics[width=0.8\linewidth]{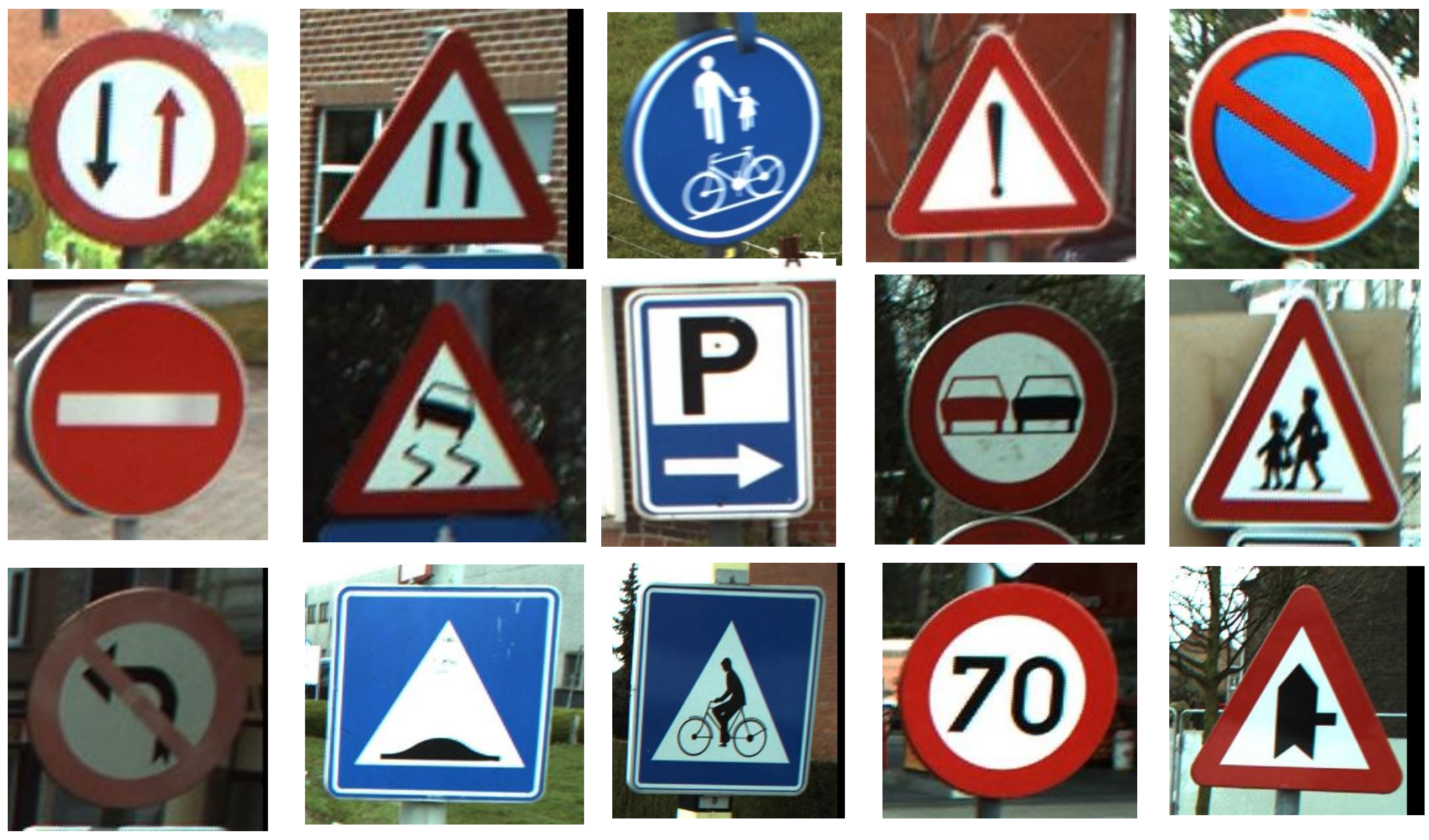}
	\caption{Overview of BelgiumTS image dataset.}
	\label{traffic}
\end{figure}

\begin{figure}[!t]
	\centering
	\includegraphics[width=0.8\linewidth]{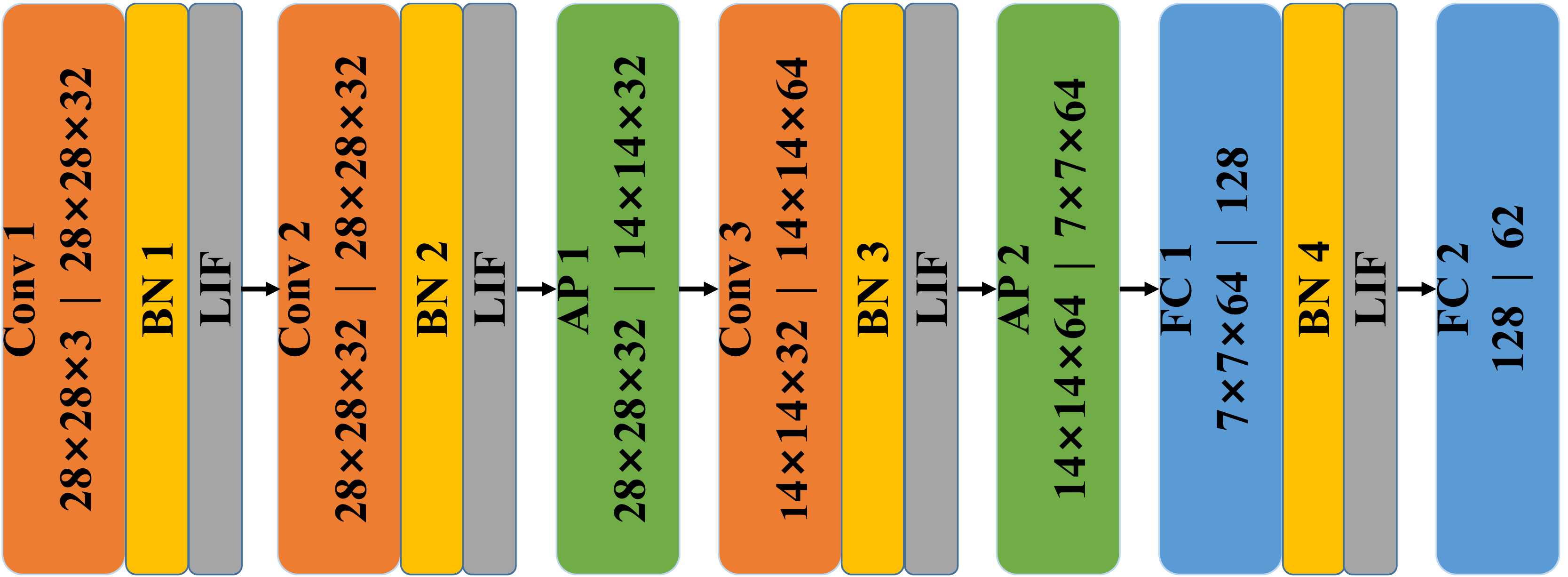}
	\caption{Description of SNN network structure. Here, Conv represents the convolution layer, AP represents the average pooling layer, and FC represents the full connection layer.}
	\label{snn}
\end{figure}

\section{Numerical Results}\label{sec-5}
In IoV, autonomous vehicles correctly perceive their surrounding environment through advanced sensors, radar, and Global Position System (GPS), and then make appropriate navigation strategies according to sensing traffic information and recognized traffic signs \cite{luo2020traffic,liu2020privacy}. Therefore, we experimentally validate our method $\mathrm{FedSNN}$-$\mathrm{NRFE}$ and baselines through traffic sign recognition tasks on a real-world dataset named Belgium traffic sign (BelgiumTS) Dataset \cite{jain2019novel}.

For all experiments, we evaluate the performance of the proposed $\mathrm{FedSNN}$-$\mathrm{NRFE}$ and baseline schemes on a desktop computer with an i9-9900K CPU and RTX 2080Ti GPU. For the federated learning tasks, we consider that there are  $K$ vehicles acting as clients and one RSU working as the parameter server to execute traffic sign recognition tasks.

\subsection{Experiment Setup}
\textbf{1) Dataset under IID and non-IID settings:} As shown in Fig. \ref{traffic}, the BelgiumTS dataset has 62 classes of 28$\times$28 RGB color traffic sign images, including 4575 training images and 2520 test images. 
For the IID setting of data distribution in FL, each client holds 186 training samples, in which there are 3 samples for each class and a total of 20 clients.
For the non-IID settings of data distribution, we simulated class imbalances in the local data distribution of different clients.
Specifically, we generate different non-IID levels by sampling the training set by adjusting parameter $\mu$ of the Dirichlet distribution function \cite{yurochkin2019bayesian,hsu2019measuring}.
The smaller the parameter $\mu$, the more obvious the class imbalance of the local training set between different clients; on the contrary, the less obvious it will be. Here, we use $\mu=0.5$ as the non-IID setting of FL.

\textbf{2) Model architecture:}
As shown in Fig. \ref{snn}, the model structure of the SNN used in our experiment consists of three convolution layers, four BN layers, two pooling layers, and two fully connected layers.
Since the SNN layer transmits a binary sequence among layers, maximum pooling will cause a large amount of information to be lost in the next layer.
Therefore, we use the average pool during training.
% Moreover, we added the BN layer after each convolutional layer.

\textbf{3) Baselines:}
Our baselines include: 1) Centralized $\mathrm{SNN}$: traditional and centralized spike neural network;
2) $\mathrm{FedCNN}$: federated convolutional neural network with the same model architecture;
3) $\mathrm{FedSNN}$: federated spike neural network coded by standard normal distribution of receptive fields (i.e., $G_{s \times s} \sim N(0, 1)$).

\textbf{4) Training details:}
In the experiment, we set the total number of clients $K=20$, the participation rate $F=0.5$, local training epoch number $E=2$,  mini-batch size $B=8$ and learning rate $\eta=1e-1$.
For the SNN, the time step is $T=10$, the leakage rate $\lambda = 0.9$, the substitution rate $\alpha=0.3$, and the threshold voltage $\theta = 1.0$.

\subsection{Experiment Results}
\subsubsection{The performance impacts of IID and non-IID setting}
\begin{figure*}[t]
	\centering
	\subfigure[]{\includegraphics[width=0.35\linewidth]{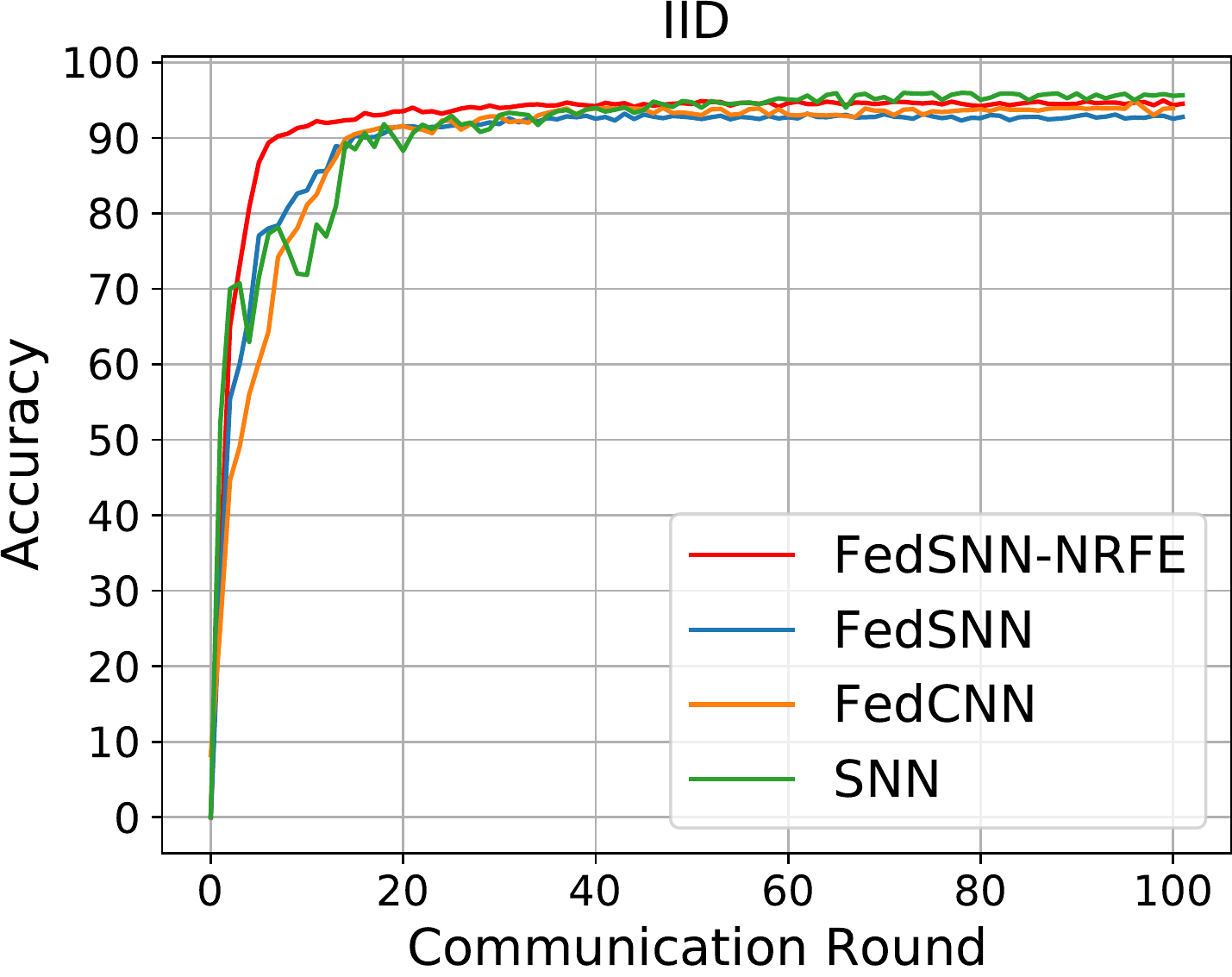}
		\label{fig-1}}
	\subfigure[]{\includegraphics[width=0.35\linewidth]{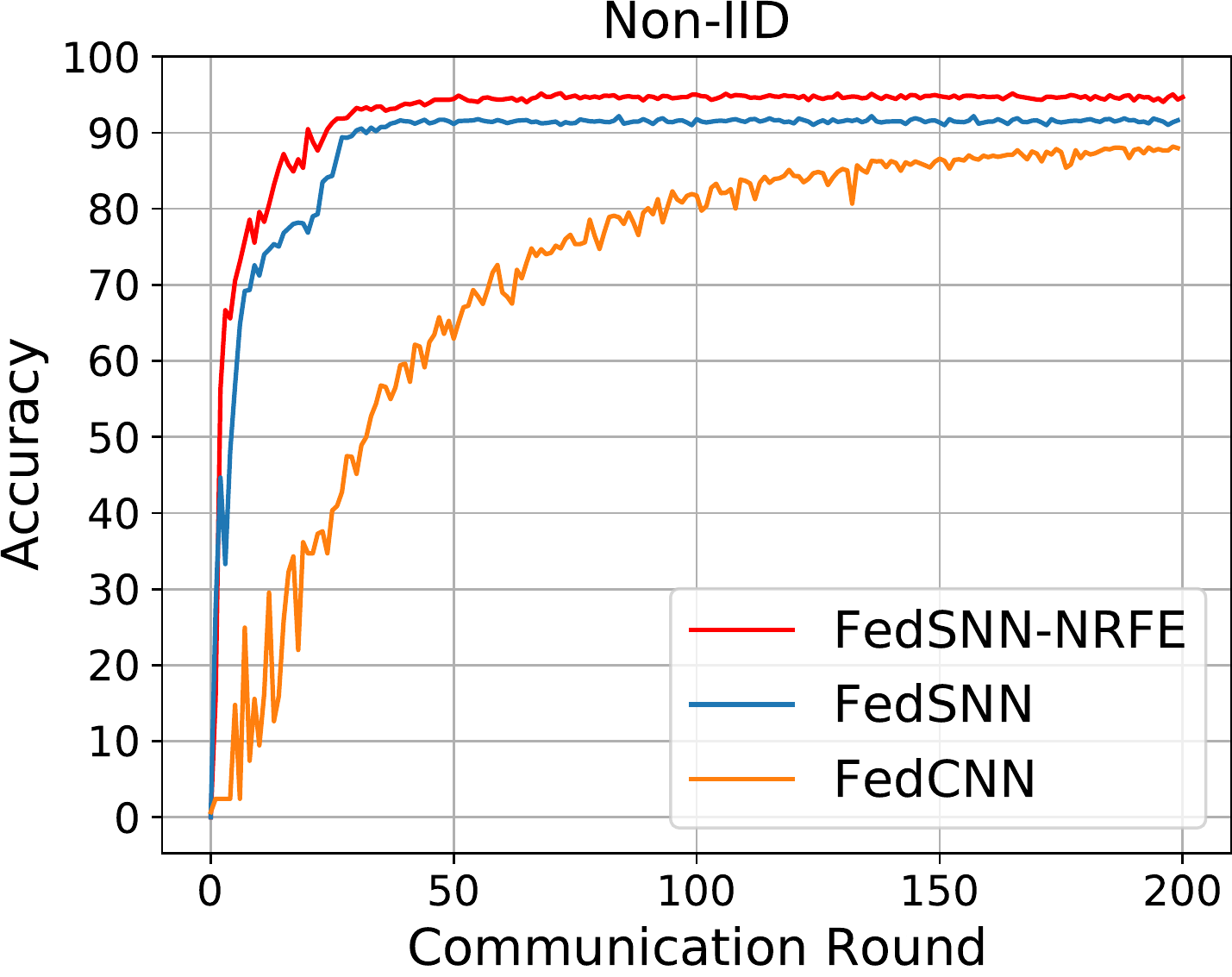}
		\label{fig-2}}
	\caption{Test accuracy curves of (a) IID  and (b) non-IID settings of data distribution.}
	\label{12}
\end{figure*}

In Table \ref{tab-1}, we show the average accuracy and fluctuation range of test results of centralized and federated learning methods under IID and non-IID settings.
Obviously, the centralized SNN achieves the best performance thus validating that the spike neural network can be well applied to the image classification tasks in IoV.
However, for IoV scenarios, it is impractical to gather all the sensing traffic signs from distributed vehicles into a centralized database because of the high risk of user privacy leakage.
Therefore, we introduce the paradigm of federated learning to effectively solve the privacy protection issues. Fig. \ref{fig-1} shows that, compared with centralized SNN, $\mathrm{FedSNN}$ and $\mathrm{FedSNN}$-$\mathrm{NRFE}$ almost achieved the same test accuracy.
Meanwhile, from Fig. \ref{fig-2}, we observe that the spike neural network has a significant advantage at the level of non-IID data distribution.
Specifically, the accuracy levels  of  $\mathrm{FedSNN}$-$\mathrm{NRFE}$ and $\mathrm{FedSNN}$ exceed that of $\mathrm{FedCNN}$ by 7\% and 4\%, respectively.
These results show that $\mathrm{FedSNN}$-$\mathrm{NRFE}$ is very effective on the BelgiumTS dataset and hence well-suited for the traffic image recognition task in IoV scenarios.

\begin{table}[!t]
	\centering
	\caption{Performance comparison of different methods under IID and non -IID settings.}
	\begin{tabular}{|c|c|c|}\hline
		\multicolumn{3}{|c|}{BelgiumTS with 20 clients (K=20, F=0.5)}\\\hline
		Methods&IID&non-IID\\\hline
		Centralized SNN&95.86$\pm$0.15&N/A\\
		FedCNN&92.89$\pm$0.23&87.75$\pm$0.36\\
		FedSNN&92.93$\pm$0.18&91.62$\pm$0.54\\
		\textbf{FedSNN-NRFE}&\textbf{94.95$\pm$0.28}&\textbf{94.88$\pm$0.36}\\\hline
	\end{tabular}
	\label{tab-1}
\end{table}

\begin{figure*}[t]
	\centering
	\subfigure[]{\includegraphics[width=0.3\linewidth]{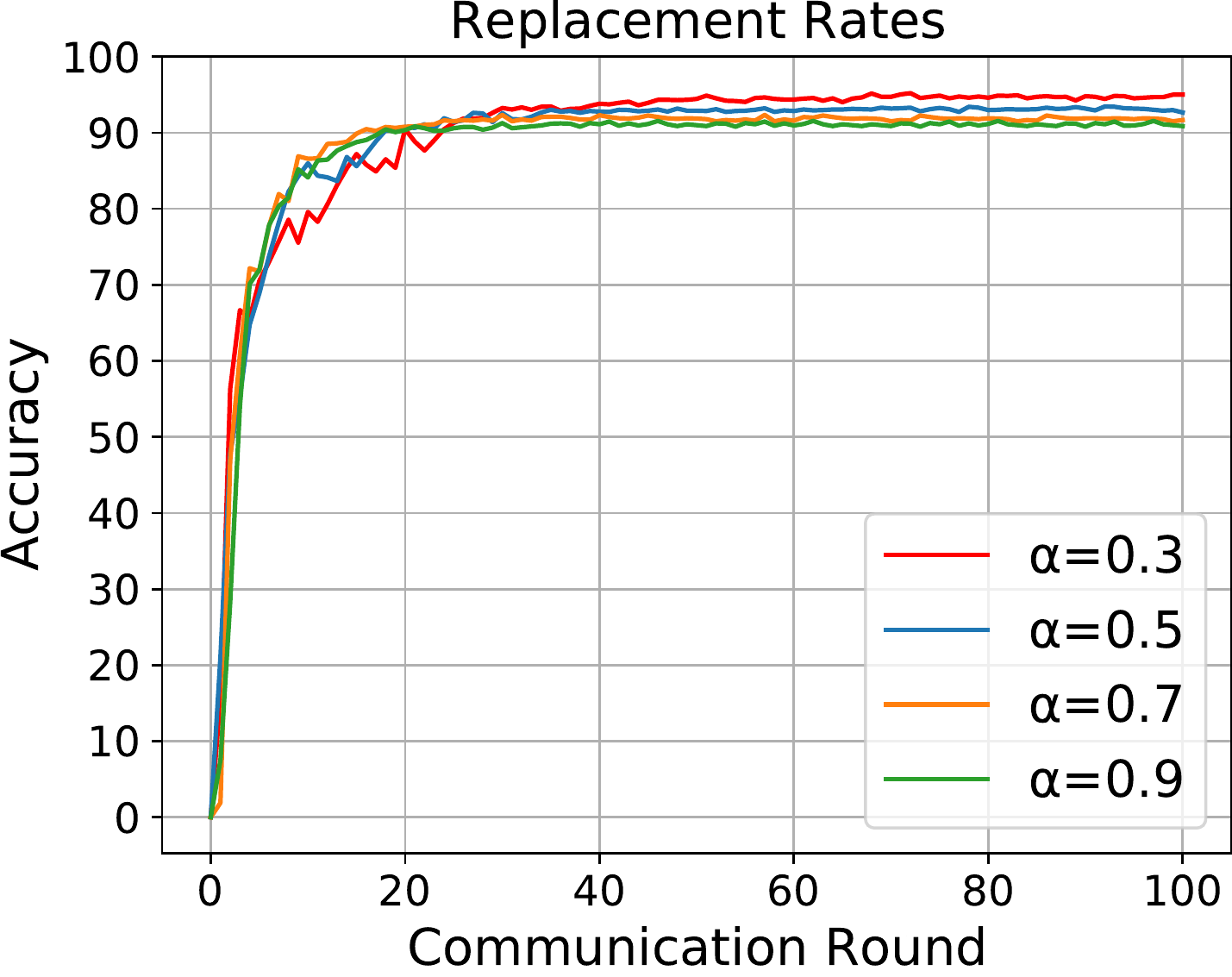}
		\label{fig-7}}
	\subfigure[]{\includegraphics[width=0.3\linewidth]{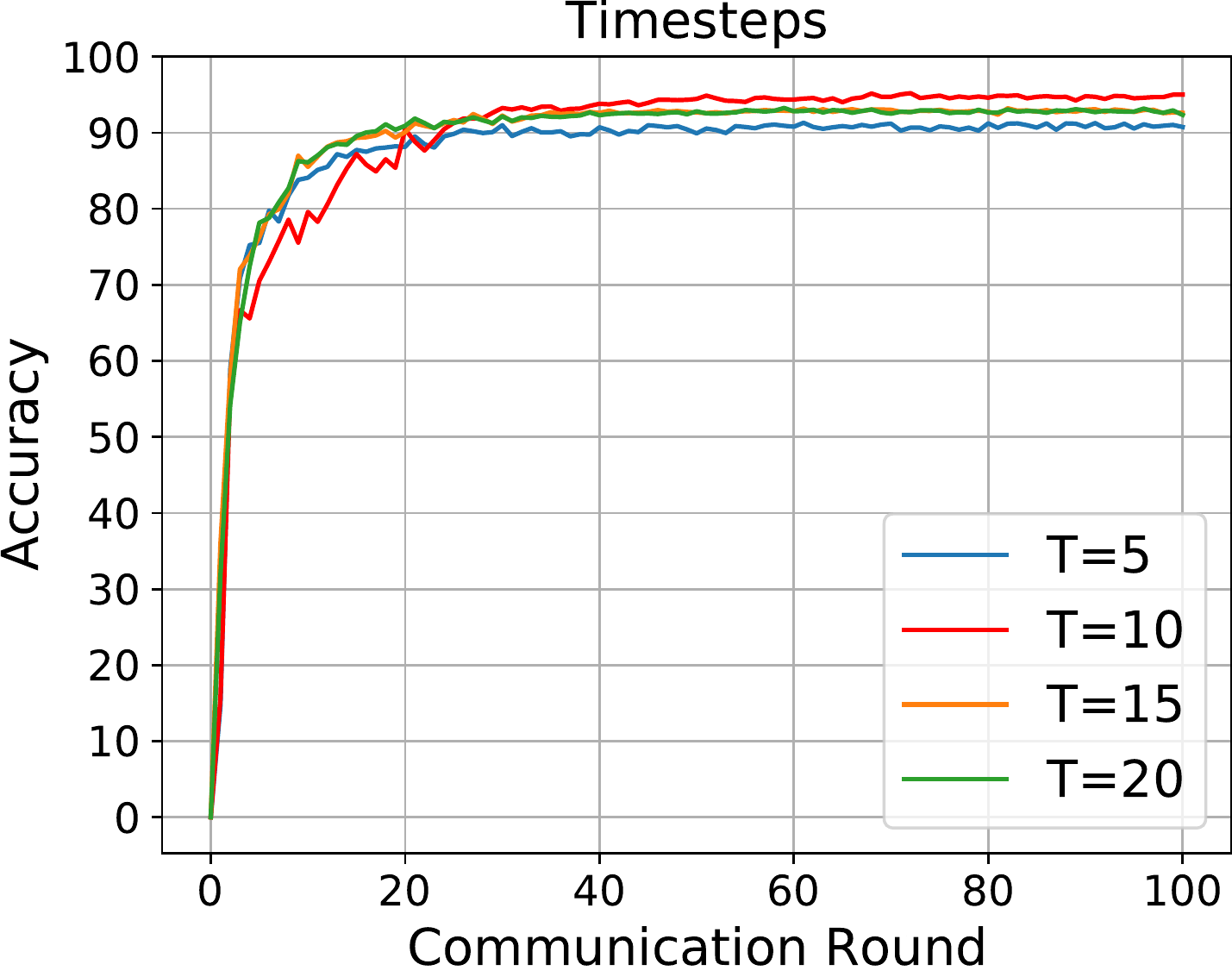}
		\label{fig-8}}
	\subfigure[]{\includegraphics[width=0.3\linewidth]{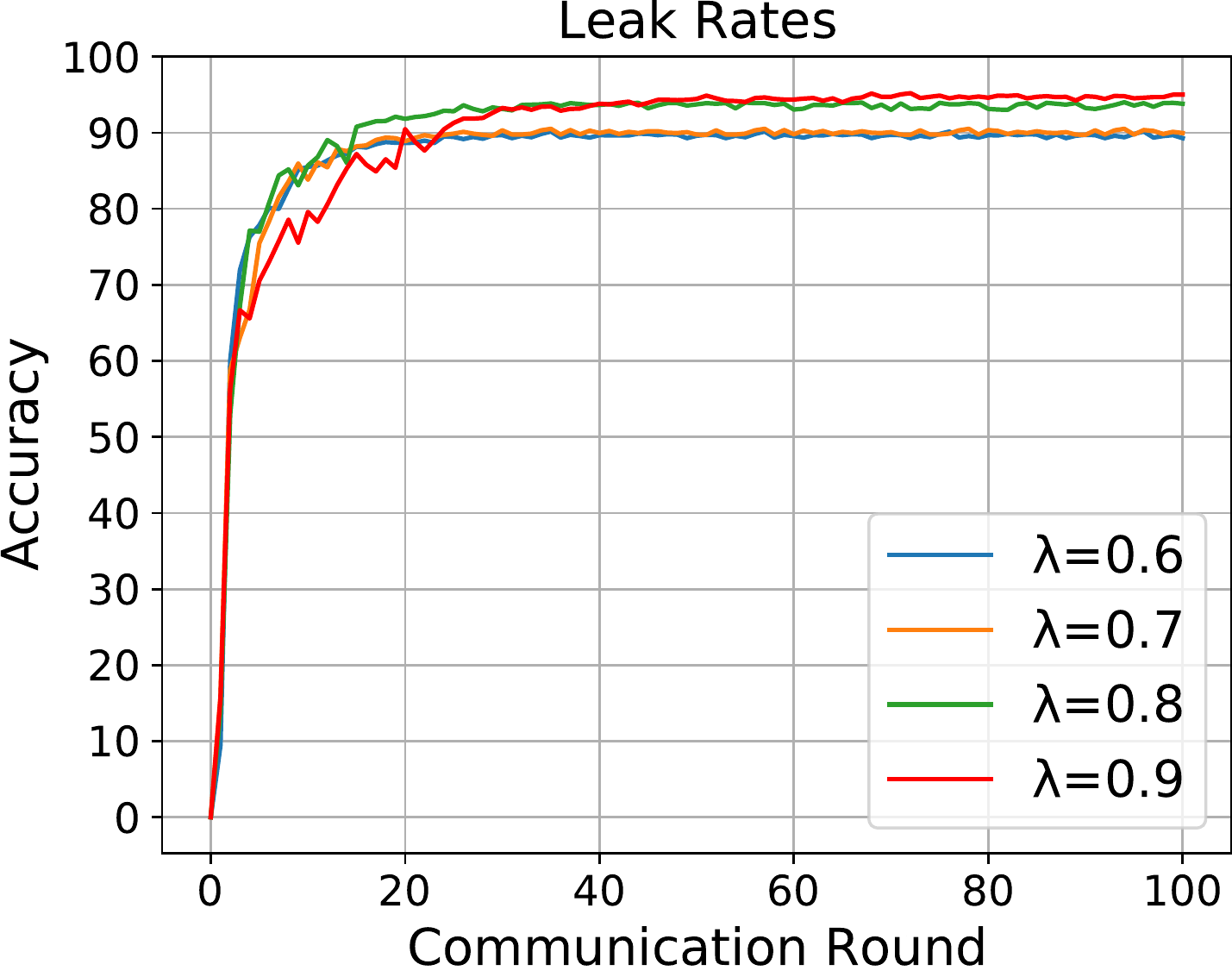}
		\label{fig-9}}
	\caption{Test accuracy curves of (a) different substitution rates, (b) different time steps, and (c) different leak rates.}
\end{figure*}

\subsubsection{Sensitivity of hyperparameters}
{To validate and analyze the optimal hyperparameter settings of the proposed $\mathrm{FedSNN}$-$\mathrm{NRFE}$, we compare the performance under different time steps, substitution rates, and leak rates.
As shown in Fig. \ref{fig-7}, the substitution rate is negatively correlated with the model performance. The larger substitution rate brings the lower test accuracy of the final convergence of the model.
Meanwhile, it is worth noting that a larger substitution rate speeds up the convergence of the model. The reason is that the large substitution rate leads to too large model updates, and it is easier to achieve the local optimum, but miss the global optimum. Conversely, the lower substitution rate makes the model update slowly and achieves higher convergence accuracy.
Therefore, in our experiments, we set the substitution rate  $\alpha$ as $0.3$.}

\begin{figure}[t] 	\centering 	\includegraphics[width=0.8\linewidth]{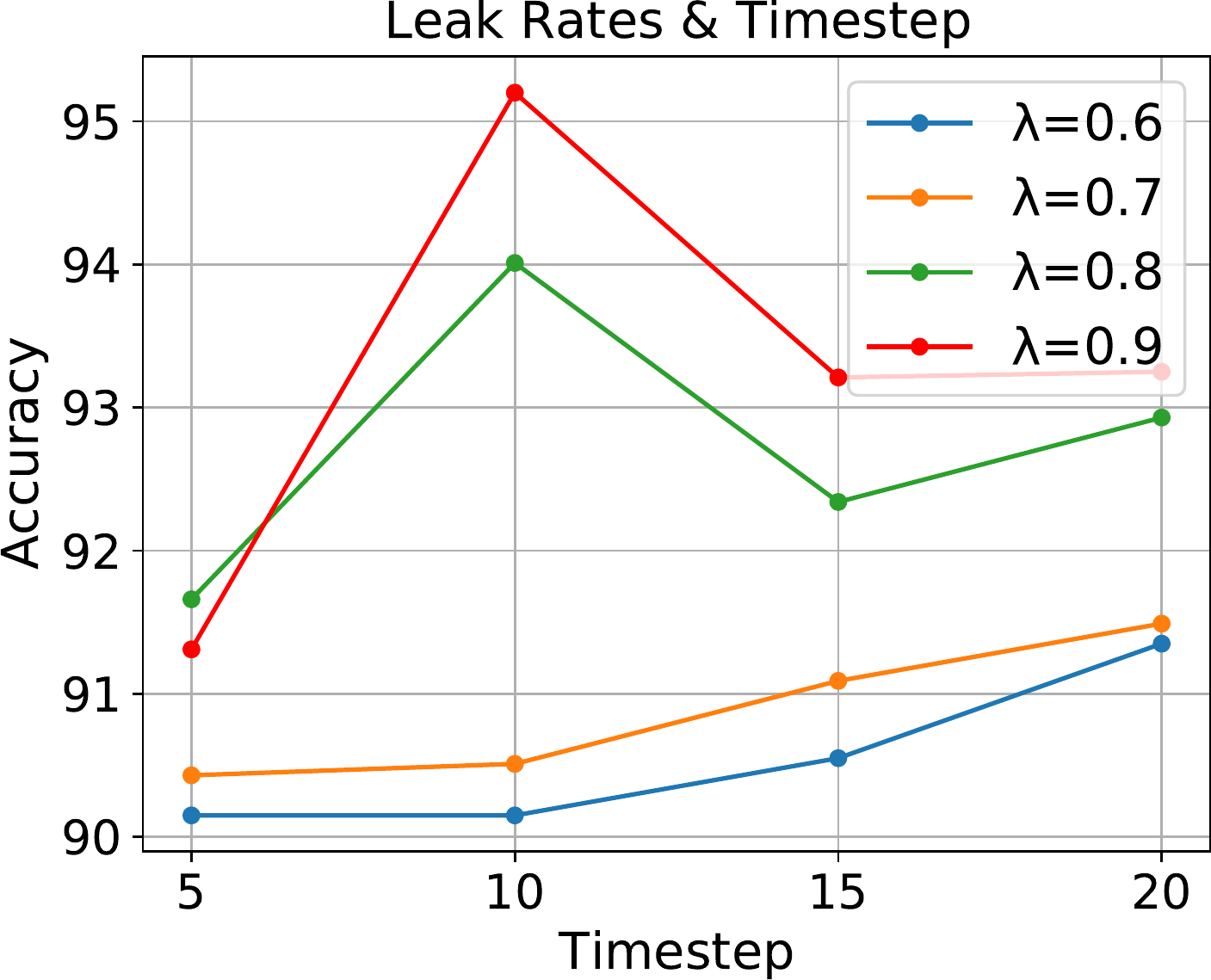} 	\caption{Performance comparison of different time steps and leak rates.} 	\label{fig-3} \end{figure}

% In Fig. \ref{fig-3}, we explored the impact of time step and leakage rate in SNN on the performance of the global model.
{For SNNs, too small a time step $T$ and leak rate $\lambda$ make it difficult to activate neurons to emit spikes, but too large a time step will cause the model to have a gradient explosion during the update process.
Therefore, it is necessary to explore the impact of these two hyperparameters on the global model performance.
As we can observe from Fig. \ref{fig-8} and Fig. \ref{fig-9}, the best test accuracy is achieved with a time step of $T=10$ and a leak rate of $\lambda=0.9$.}
Meanwhile, in Fig. \ref{fig-3}, regardless of the value of the time step $t$, the performance of leakage rate $\lambda=0.9$ is significantly or approximately better than that of the leakage rate $\lambda=0.8$.
This is because the leakage rate is positively correlated with the probability of neurons being activated.
In detail, the larger leakage rate of the spiking neuron brings the more membrane voltage it retains at the previous moment, and the easier it to activate the neuron to emit the spike.
Moreover, the setting of time steps $T$ is also important for SNNs.
It is difficult for the membrane potential of spike neurons to exceed the threshold voltage if the time step is too small. 
An excessively large time step increases the computational complexity and is prone to gradient explosion and over-fitting problems.
Through experimental verification, we observe that the time step $T=10$ and the leak rate $\lambda=0.9$ are the optimal settings for the proposed $\mathrm{FedSNN}$-$\mathrm{NRFE}$.

\begin{figure*}[t]
	\centering
	\subfigure[]{\includegraphics[width=0.3\linewidth]{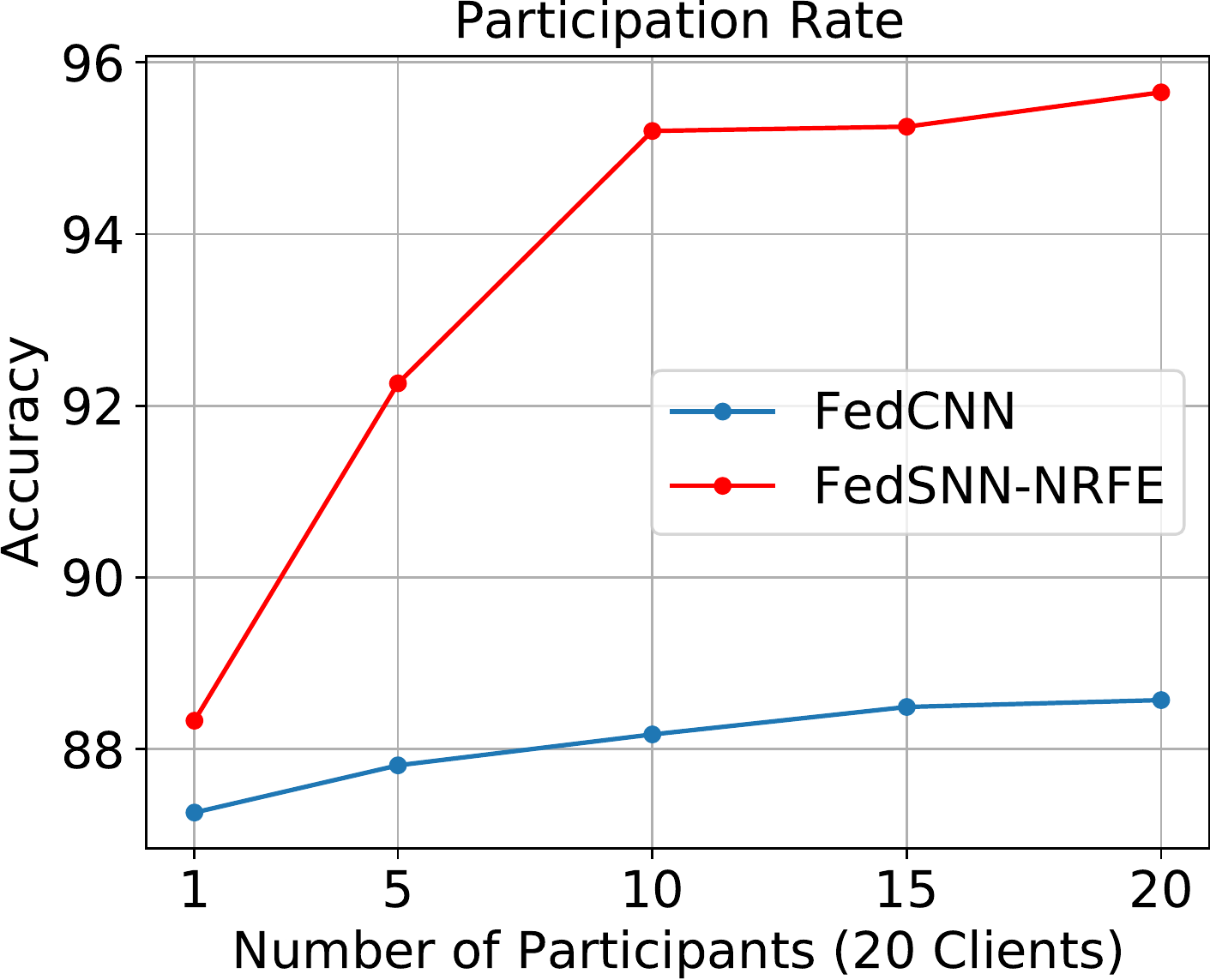}
		\label{fig-12}}
	\subfigure[]{\includegraphics[width=0.3\linewidth]{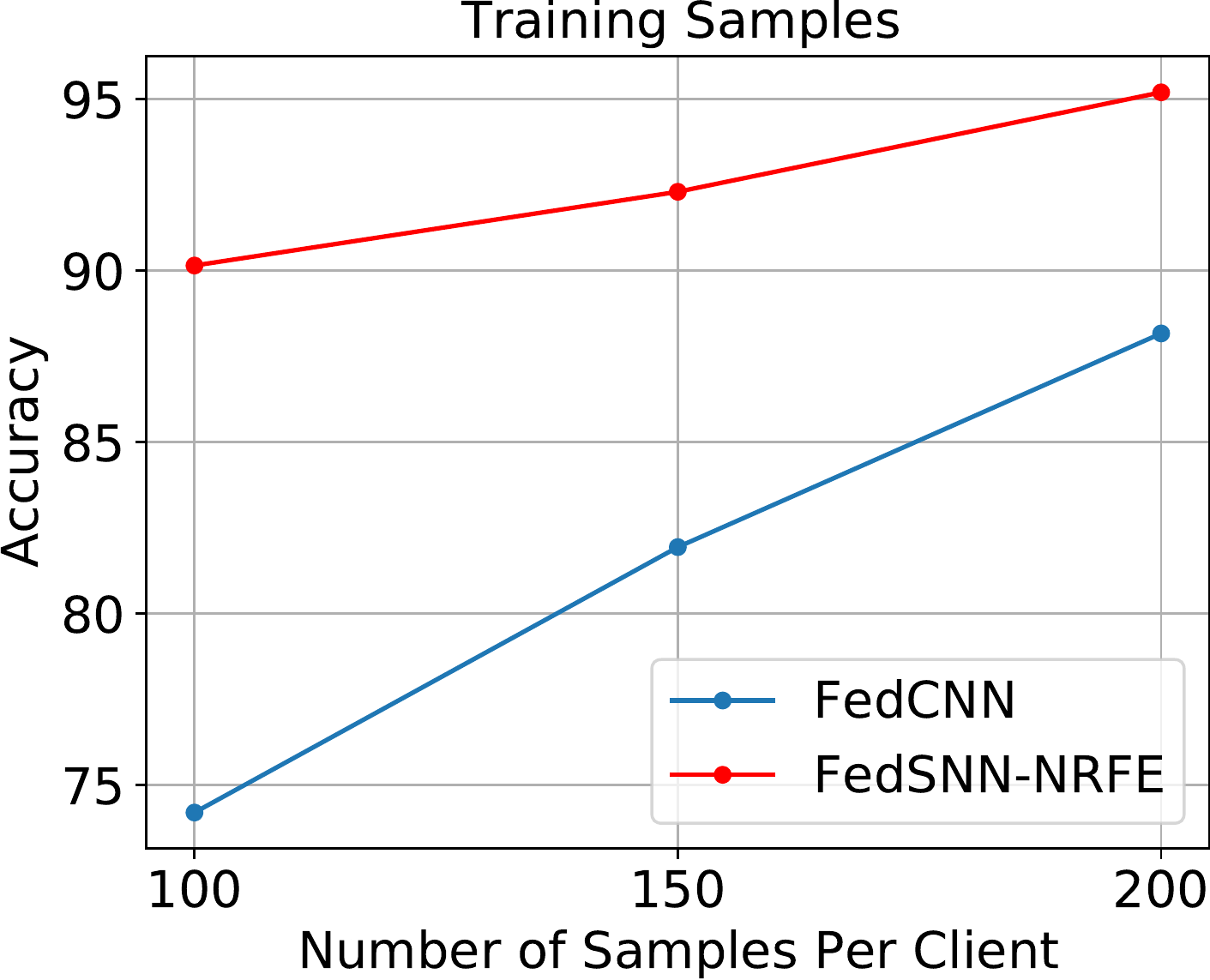}
		\label{fig-4}}
	\subfigure[]{\includegraphics[width=0.3\linewidth]{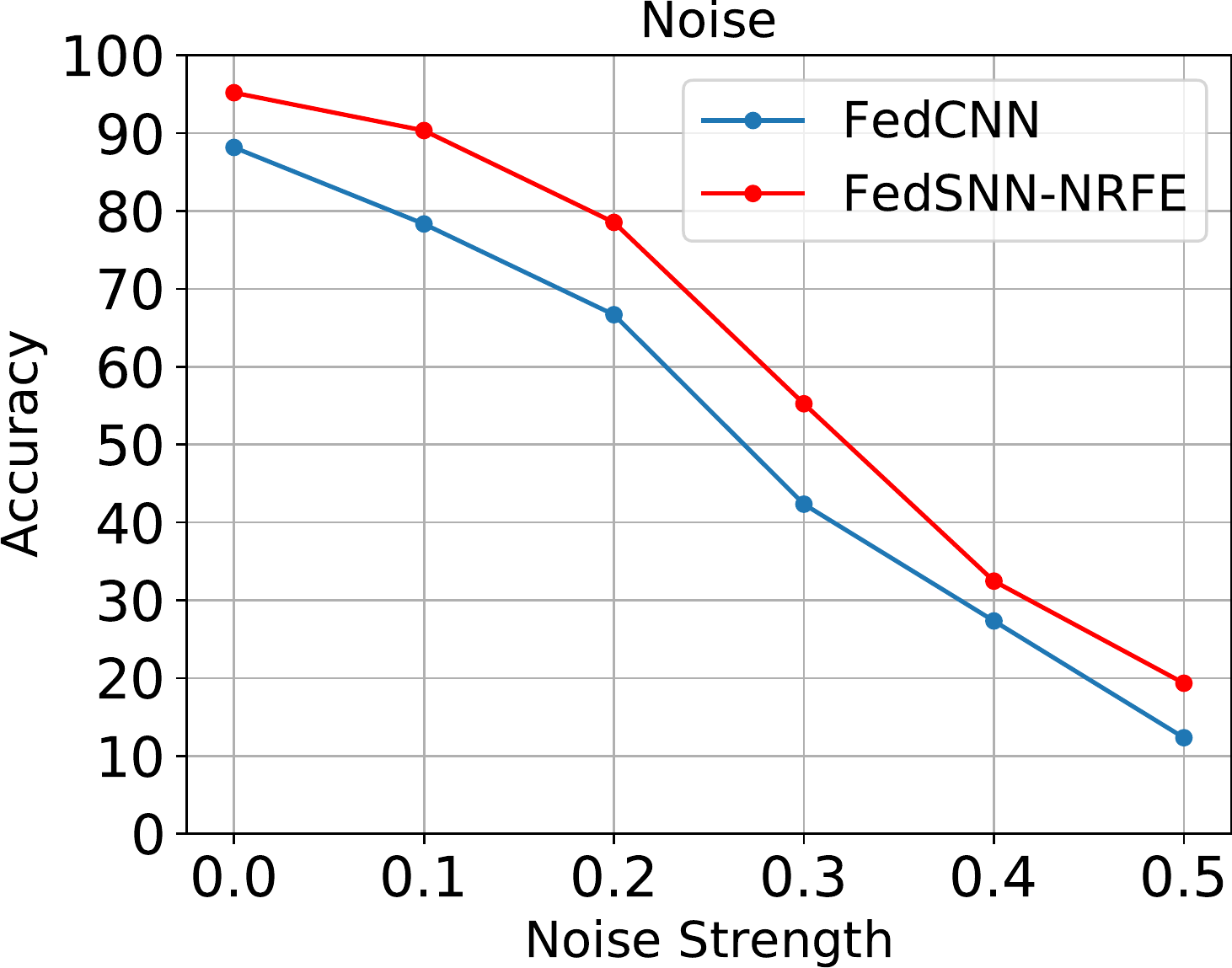}
		\label{fig-5}}
	\caption{Performance comparison of (a) different of participating clients, (b) different training samples, and (c) different noise strengths.}
	\label{123}
\end{figure*}

\subsubsection{The performance impacts on the number of participating clients}
{In FL, there are a large number of dropped devices due to system heterogeneity factors such as network connection (3G, 4G, 5G, WiFi), hardware (CPU, RAM), battery capacity, etc.
To reduce computational complexity and communication overhead, we prefer to obtain a high-quality global model in the setting of fewer clients participating in each training round.
To this end, in the non-IID case, we explore the performance impact of different numbers of clients participating in training.
Fig. \ref{fig-12} presents the maximum  accuracy level converged by our method and the baseline $\mathrm{FedCNN}$ after 200 rounds with the participation rate $F = 5\%, 25\%, 50\%, 75\%, 100\%$ settings.
Obviously, as the number of participating clients increases, the performance also increases gradually.
Especially for $\mathrm{FedSNN}$-$\mathrm{NRFE}$, when the number of participations is increased from 1 to 10, the performance improvement is significant and much higher than the baseline $\mathrm{FedCNN}$.
This implies that it is feasible to integrate SNNs into FL.}
\subsubsection{The performance impacts of the number of training sets}
In traditional ANNs, a high-precision traffic sign recognition model usually requires a large number of training samples.
{However, this increases the cost of collecting data and the time complexity of training the model to a certain extent.
SNNs are the third generation of artificial neural networks inspired by biological information processing, which have the advantages of fast reasoning and event-driven information processing.
To this end, we compare the performance of CNN and SNN for traffic sign image recognition under the FL paradigm.}
Fig. \ref{fig-4} shows the performance comparison of 200, 150, and 100 training samples per client under non-IID settings.
Obviously, for $\mathrm{FedCNN}$, as the number of local training samples held by the client decreases, the test accuracy of the global model drops by 15\%.
In contrast to our $\mathrm{FedSNN}$-$\mathrm{NRFE}$, when the number of training samples is reduced from 200 to 100, the accuracy is only reduced by 5\%.
Meanwhile, we observe that the accuracy of our method is better than that of the baseline (i.e., FedCNN) when the number of training samples is only 100.
This proves that the SNN has the characteristics of strong adaptability under the condition of few training samples.

\subsubsection{The robustness performance to image noise and different non-IID data distributions}
In the real world, due to external environmental factors (e.g., sun exposure, rain erosion, weathering, etc.), traffic signs will become blurred.
% Furthermore, existing attack methods \cite{wang2019beyond, wei2020framework} can infer local data and its attributes from model parameters.
Thus, to simulate this phenomenon, we add different intensities of salt-and-pepper noise \cite{chan2005salt} to the BelgiumTS training dataset to verify the robustness of our $\mathrm{FedSNN}$-$\mathrm{NRFE}$ model.
Specifically, for an image, we randomly change the pixel value of a fixed proportion from the original to 0 or 255.
The observation result is shown in Fig. \ref{fig-5}.
As the proportion of added noise increases, the performance of $\mathrm{FedSNN}$-$\mathrm{NRFE}$ and $\mathrm{FedCNN}$ decreases significantly.
We observe that when the noise proportion increased from 0.2 to 0.4, the performance of both our method and baseline decreased by more than 35\%.
It is worth noting that under the same noise proportion, the test accuracy of $\mathrm{FedSNN}$-$\mathrm{NRFE}$ is better than that of $\mathrm{FedCNN}$.
Furthermore, when the salt-and-pepper noise ratio is 0.1, the performance of our proposed method still reaches 91\%.
In short, compared to CNNs, SNNs are more robust to noisy images in IoV.

\begin{figure}[t]
	\centering
	\includegraphics[width=0.8\linewidth]{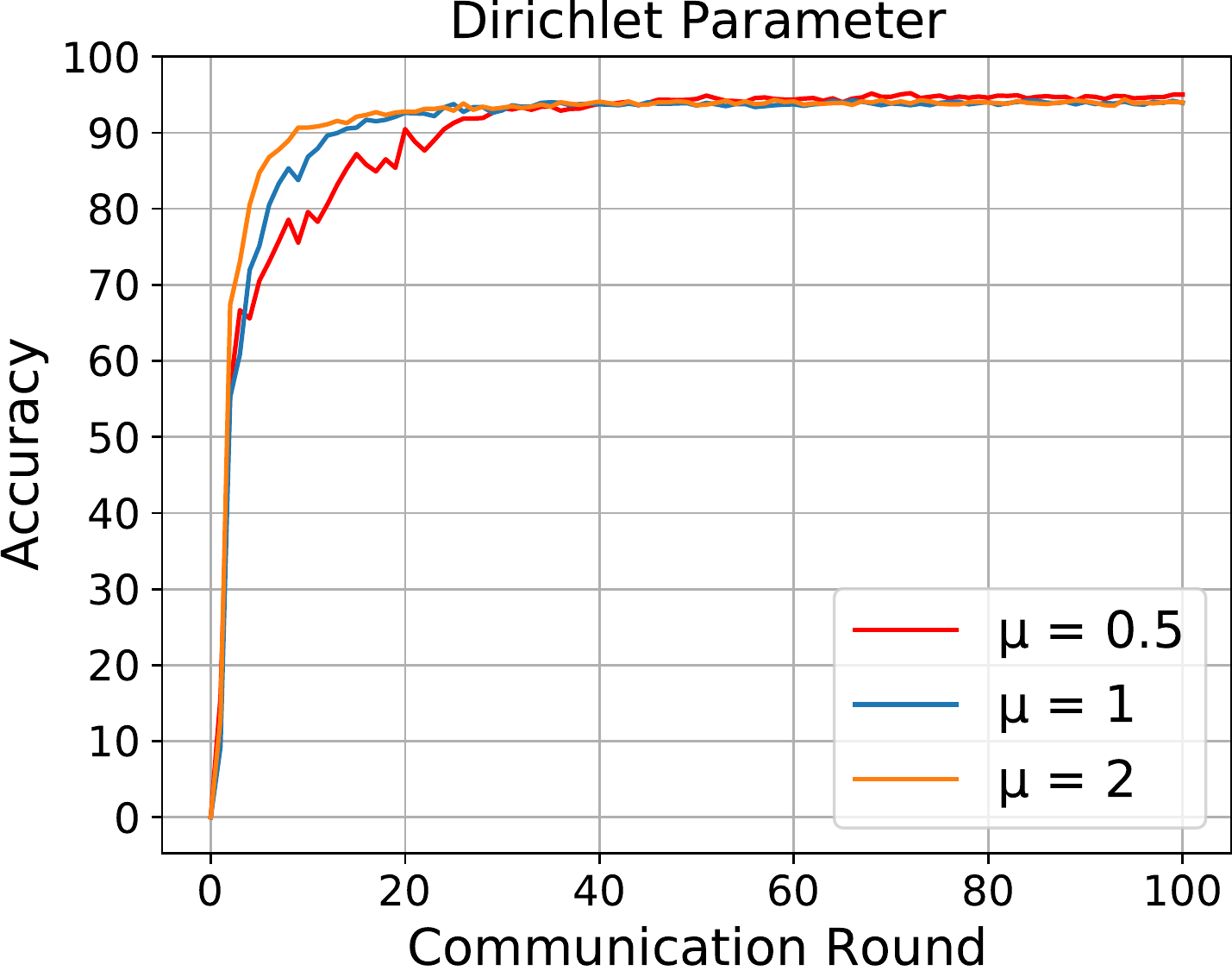}
	\caption{Test curves for different non-IID levels.}
	\label{fig-11}
\end{figure}

{In the FL-based IoV scenarios, the traffic signs captured by vehicles using different types of in-vehicle cameras in different areas usually have large differences in features and categories.
Therefore, the data distribution among vehicles is non-IID.
Fig.~\ref{fig-11} shows the test accuracy of $\mathrm{FedSNN}$-$\mathrm{NRFE}$ on traffic sign data with different non-IID levels of vehicles.
The difference is that the higher non-IID level results in the slower convergence of the global model.
Interestingly, our proposed method achieves stable performance at all three non-IID levels and differs very little from each other.}

\begin{figure*}[!t]
	\centering
	\includegraphics[width=0.65\linewidth]{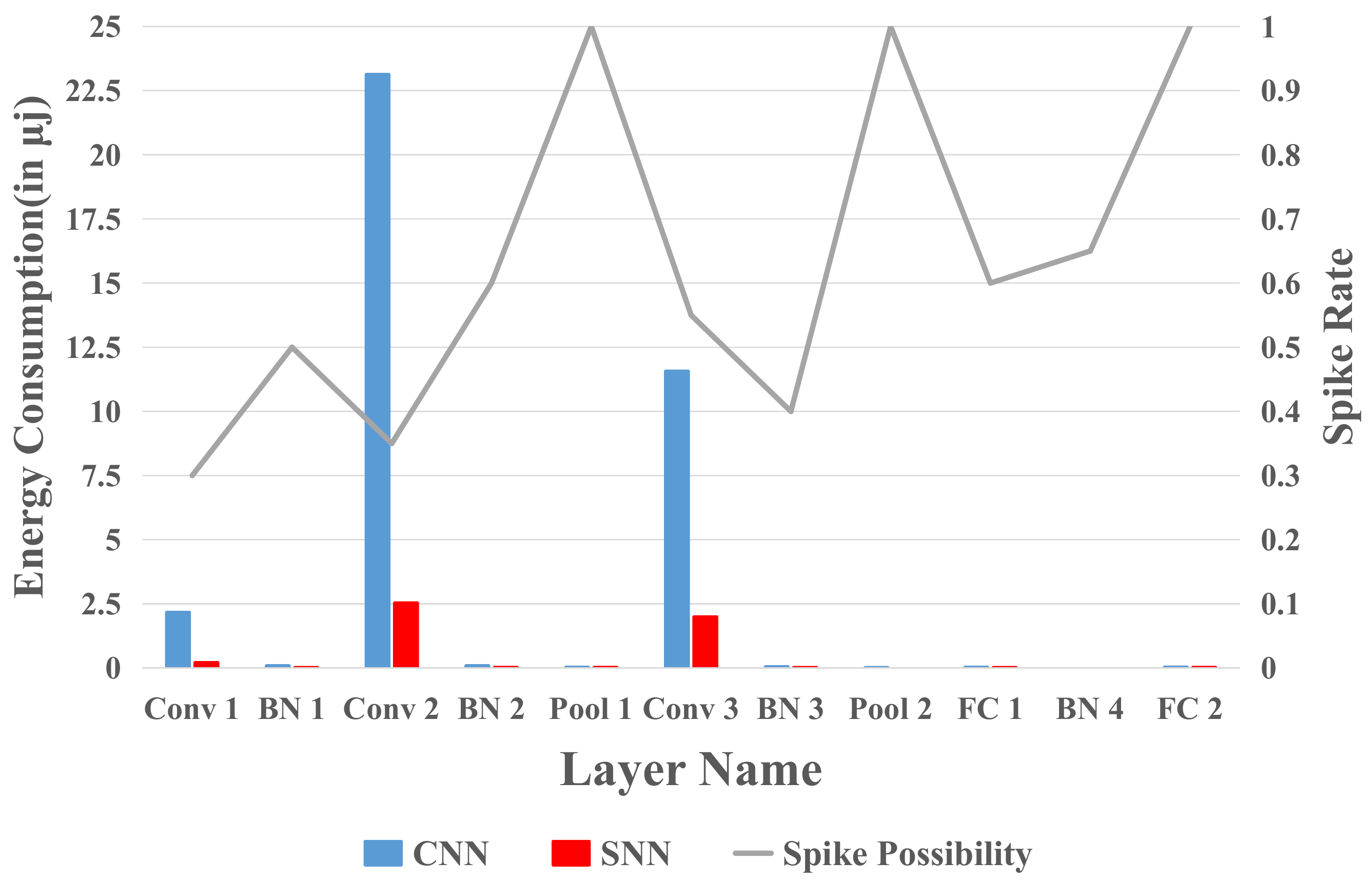}
	\caption{Comparison of the energy consumption of each layer of SNN and CNN after 100 rounds of communication.}
	\label{fig-6}
\end{figure*}

\subsubsection{The performance of  energy consumption}
To better evaluate the computational cost of CNNs and SNNs during the training process, we use the energy consumption metric described in \cite{6757323}.
Table \ref{tab-n} shows the energy consumption of 32-bit integer arithmetic operations based on the 45nm CMOS process.
In the propagation process of neural networks, the computational cost is proportional to the calculation of floating-point numbers.
In our experiment, we only consider multiplication and addition operations.
Specifically, for layer $l$ in CNNs, the number of floating-point operations (FLOPs) is defined as:
\begin{equation}
FLOPs(l) = \begin{cases}ks^{2} \times M_{out}^{2} \times C_{in} \times C_{out}, &Conv
\\C_{in} \times M_{in}^{2}, &BN\ or AP
% \\C_{in} \times H_{out} \times W_{out}&AP
\\N_{in} \times N_{out}, &{FC}\end{cases},
\end{equation}
where $ks$ is kernel size. $C_{in}$ is input channel and $C_{out}$ is output channel. {$M_{in}$ and $M_{out}$ are the side lengths of the input and output feature maps.}
$N_{in}$ is input neuron numbers and $N_{out}$ is output neuron numbers.

For SNNs, floating-point operations are only generated when the neuron outputs pulses.
Meanwhile, binary propagation performs is an accumulation (AC) operation, resulting in significant energy efficiency \cite{9583900}.
We define the spike rate $R_{s}$ of each layer, i.e., the number of spikes $n_{l}^{s}$ sent by neurons in the $l$-th layer within the time step $T$ divided by the number of neurons $N_{l}$ in the $l$-th layer. 
The calculation formula of spike rate $R_{s} (l)$ of the $l$-th layer is given as follows:
\begin{equation}
    R_{s}(l)=\frac{n_{l}^{s}}{N_{l}}.
\end{equation}
% Therefore, the number of floating-point operations in the $l$-th layer in SNN is:
% \begin{equation}
%     FLOPs_{SNN}(l)=FLOPs(l) \times R_{s}(l).
% \end{equation}
Next, we calculate the energy consumption of each layer of CNN and SNN separately:
\begin{equation}
E_{CNN}(l) =FLOPs(l) \times (E_{Mult} + E_{Add}),
\end{equation}
\begin{equation}
\begin{aligned}
E_{SNN}(l) =FLOPs(l) \times R_{s}(l) \times T \times E_{AC}.
\end{aligned}
\end{equation}
Because a larger time step will inevitably lead to higher computational complexity, we set the time step $T=10$ in our experiment.
Fig. \ref{fig-6} shows the estimated energy consumption of each layer of CNNs and SNNs for the model of a client trained on the BelgiumTS dataset.
Intuitively, the energy consumption of floating-point operations is mainly concentrated in the convolutional layer.
Compared with CNNs, we observe that the energy consumption of SNNs mainly depends on the ratio of activated neurons in each layer of the network.
Table \ref{tab-5} provides the details about the energy consumption and spike rate of each layer.
The total energy estimated by CNN is about \textbf{37.096 $\mu$J}  while that of SNN is about \textbf{4.773 $\mu$J}, which is  \textbf{7.772$\times$} more efficient than of CNN.

\begin{table}[!t]
	\centering
	\caption{Estimate the energy of each multiplication and addition operation.}
	\begin{tabular}{|c|c|}\hline
		\textbf{Operation} & \textbf{Estimated Energy (pJ)}\\\hline
		32-bit Multiply $(E_{Mult})$ & 3.1\\
		32-bit Add $(E_{Add})$ & 0.1\\
		32-bit Accumulate $(E_{AC})$ & 0.1\\\hline
	\end{tabular}
	\label{tab-n}
\end{table}

\begin{table}[!t]
	\centering
	\caption{Comparison of the energy consumption of each layer of CNN and SNN.}
	\begin{tabular}{|c|c|c|c|}\hline
		\textbf{Layer Name} & \textbf{CNN (pJ)} & \textbf{SNN (pJ)} & \textbf{Spike Rate}\\\hline
		Conv 1 &2,167,603.2 &203,212.8 &0.3\\
		BN 1 &80,281.6 &12,544 &0.5 \\
		Conv 2 &23,121,100.8 &2,528,870.4 &0.35\\
		BN 2 &80,281.6 &15,052.8 &0.6 \\
		AP 1 &21,952 &6,860 &1.0 \\
		Conv 3 &11,560,550.4 &1,986,969.6 &0.55\\
		BN 3 &40,140.8 &5,017.6 &0.4 \\
		AP 2 &10,579.2 &3,306 &1.0 \\
		FC 1 &20,070.4 &3,763.2 &0.6 \\
		BN 4 &409.6 &83.2 &0.65 \\
		FC 2 &25,395.2 &7,936 &1.0 \\\hline
	\end{tabular}
	\label{tab-5}
\end{table}

\section{Conclusion}\label{sec-6}
\textcolor{black}{In this paper, we conduct a comprehensive study on the feasibility of training spike neural networks with the federated learning paradigm in the Internet of Vehicles scenarios.
Specifically, we propose a $\mathrm{FedSNN}$-$\mathrm{NRFE}$ scheme based on the neuronal receptive field encoding, which realizes the high-precision recognition of road traffic signs in autonomous driving.
Furthermore, as far as we know, we are the first to train FedSNN under the traffic sign dataset, and conduct a comprehensive analysis of performance, adaptability, robustness, and energy efficiency.
Through numerical results, we conclude that in the federated learning scenario of the Internet of Vehicles, SNN has become a viable alternative to CNN with its superior energy efficiency and noise resistance.
Nevertheless, FL still faces challenges in terms of expensive communication overhead, unreliable devices, and security.
There are several possible directions that are worth being studied: I) Due to the binary spike sequences passed between neurons in SNNs, it is promising to improve the aggregation algorithm of FL for efficient communication overhead.
II) Considering that client selection strategies based on SNN-related metrics (such as spike rate) should be designed to improve the performance and convergence speed of FL.
III) It is still necessary to strengthen the security of the $\mathrm{FedSNN}$-$\mathrm{NRFE}$ training algorithm to ensure strict confidentiality of user privacy data.
In future work, we will further compare and evaluate these methods for integrating SNNs into FL.}
\bibliographystyle{IEEEtran}
\bibliography{ref}

\begin{IEEEbiography}[{\includegraphics[width=1in,height=1.25in,clip,keepaspectratio]{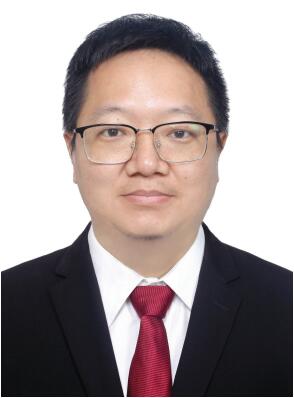}}]{Kan Xie}
received the Ph.D. degree in Control Science and Engineering from the Guangdong University of Technology, Guangzhou, China, in 2017. He joined the Institute of Intelligent Information Processing, Guangdong University of Technology, where he is currently an Associate Professor. His research interests include machine learning, non-negative signal processing, blind signal processing, smart grid, and Internet of Things.
\end{IEEEbiography}

\begin{IEEEbiography}[{\includegraphics[width=1in,height=1.25in,clip,keepaspectratio]{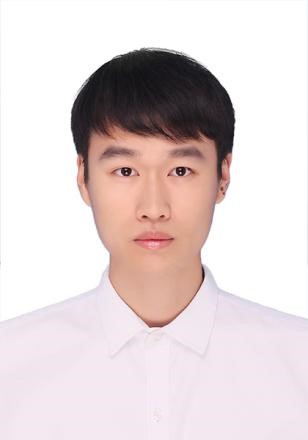}}]{Zhe Zhang}
received the B.S. degree from the North China University of Water Resources and Electric Power, in 2020. He is currently working toward a Master's Degree in Information Statistics Technology at Heilongjiang University. His research interests are mainly federated learning, semi-supervised learning, and spiking neural networks.
\end{IEEEbiography}

\begin{IEEEbiography}[{\includegraphics[width=1in,height=1.25in,clip,keepaspectratio]{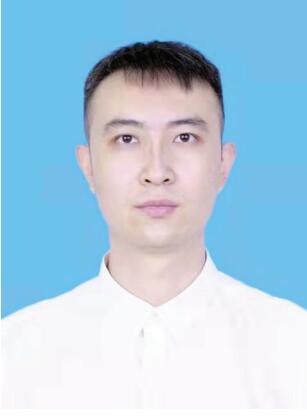}}]{Bo Li}
received the Ph.D. degree in Computer Science and Technology from the School of Intelligent Systems Engineering at Sun Yat-sen University, Guangzhou, China, in 2021. From September 2014 to June 2016, he was a master student in the school of engineering, at Sun Yat-sen University. He is currently a Post-Doctoral Research Fellow with the School of Automation, Guangdong University of Technology, Guangzhou, China. His current research interests include intelligent transportation systems, traffic information processing as well as big data technology. Dr. Li is currently a very active reviewer for some international journals and a member of the program committee for many international conferences.
\end{IEEEbiography}

\begin{IEEEbiography}[{\includegraphics[width=1in,height=1.25in,clip,keepaspectratio]{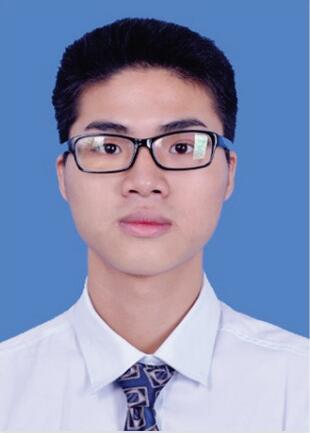}}]{Jiawen Kang}
received the Ph.D. degree from the Guangdong University of Technology, China in 2018. He has been a postdoc at Nanyang Technological University, Singapore from 2018 to 2021. He currently is a full professor at Guangdong University of Technology, China. His research interests mainly focus on blockchain, security, and privacy protection in wireless communications and networking.
\end{IEEEbiography}

\begin{IEEEbiography}[{\includegraphics[width=1in,height=1.25in,clip,keepaspectratio]{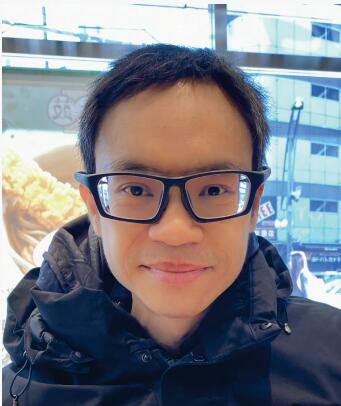}}]{Dusit Niyato (M’09-SM’15-F’17)}
is currently a professor in the School of Computer Science and Engineering, at Nanyang Technological University, Singapore. He received B.Eng. from King Mongkut's Institute of Technology Ladkrabang (KMITL), Thailand in 1999 and the Ph.D. in Electrical and Computer Engineering from the University of Manitoba, Canada in 2008. His research interests are in the areas of the Internet of Things (IoT), machine learning, and incentive mechanism design.
\end{IEEEbiography}

\begin{IEEEbiography}[{\includegraphics[width=1in,height=1.25in,clip,keepaspectratio]{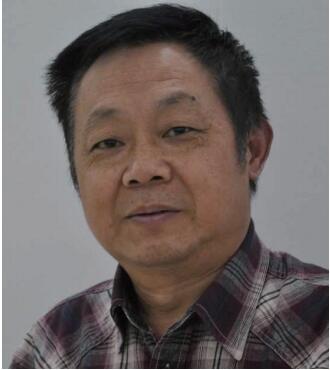}}]{Shengli Xie  (M’01, SM’02, F’19)}
received the B.S. degree in Mathematics from Jilin University, China in 1983, the M.S. degree in Mathematics from Central China Normal University, China in 1995, and the Ph.D. degree in Control Theory and Applications from South China University of Technology, China in 1997. He is currently a Full Professor and the head of the Institute of Intelligent Information Processing, Guangdong University of Technology. He has co-authored two books and over 150 research papers in refereed journals and conference proceedings and was awarded Highly Cited Researcher in 2020. His research interests include blind signal processing, machine learning, and the Internet of Things. He was awarded the Second Prize of the National Natural Science Award of China in 2009. He is an Associate Editor of IEEE TSMC.
\end{IEEEbiography}

\begin{IEEEbiography}[{\includegraphics[width=1in,height=1.25in,clip,keepaspectratio]{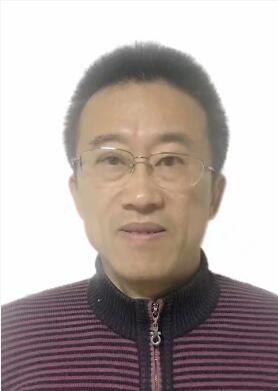}}]{Yi Wu}
received the B.S. and M.S. degrees in computer science from Heilongjiang University, Harbin, China, in 1985 and 2005, respectively. He is a Professor at the School of Data Science and Technology, Heilongjiang University.
His current research interests include large-scale data processing, data mining, and database application.
\end{IEEEbiography}

\end{document}